\documentclass[10pt]{article} 
\usepackage[accepted]{tmlr}

\usepackage{amsmath,amsfonts,bm}

\def\eqref#1{equation~\ref{#1}}

\def\1{\bm{1}}

\DeclareMathAlphabet{\mathsfit}{\encodingdefault}{\sfdefault}{m}{sl}
\SetMathAlphabet{\mathsfit}{bold}{\encodingdefault}{\sfdefault}{bx}{n}

\usepackage[colorlinks, linkcolor=black, citecolor=black, urlcolor=black]{hyperref}
\usepackage{url}
\usepackage{makecell}
\usepackage{arydshln}
\usepackage{booktabs}
\usepackage{tikz}
\usepackage{forest}
\usepackage{pifont}
\usepackage[normalem]{ulem}

\usepackage[most]{tcolorbox}
\usepackage{xcolor}
\usepackage{lipsum} 

\newcommand{\revise}[1]{\textcolor{black}{#1}}

\tcbset{
    deepseekbox/.style={colback=blue!5!white, colframe=blue!75!black, fonttitle=\bfseries, title=Answer (DeepSeek R1)},
    expectedbox/.style={colback=green!5!white, colframe=green!75!black, fonttitle=\bfseries, title=Answer (Expected)},
}

\title{Efficient Reasoning Models: A Survey}

\author{\name Sicheng Feng \email sicheng@mail.nankai.edu.cn \\
      National University of Singapore, Singapore 
      \\ Nankai University, Tianjin, China 
      \AND 
      \name Gongfan Fang \email gongfan@u.nus.edu\\
      National University of Singapore, Singapore
      \AND 
      \name Xinyin Ma \email maxinyin@u.nus.edu\\
      National University of Singapore, Singapore
      \AND 
      \name Xinchao Wang\thanks{Corresponding author} \email xinchao@nus.edu.sg\\
      National University of Singapore, Singapore
      }

\def\month{09}  
\def\year{2025} 
\def\openreview{\url{https://openreview.net/forum?id=sySqlxj8EB}} 

\begin{document}

\maketitle

\begin{abstract}
Reasoning models have demonstrated remarkable progress in solving complex and logic-intensive tasks by generating extended Chain-of-Thoughts (CoTs) prior to arriving at a final answer. Yet, the emergence of this ``slow-thinking'' paradigm, with numerous tokens generated in sequence, inevitably introduces substantial computational overhead. To this end, it highlights an urgent need for effective acceleration. This survey aims to provide a comprehensive overview of recent advances in efficient reasoning. It categorizes existing works into three key directions: (1) \emph{shorter} – compressing lengthy CoTs into concise yet effective reasoning chains; (2) \emph{smaller} – developing compact language models with strong reasoning capabilities through techniques such as knowledge distillation, other model compression techniques, and reinforcement learning; and (3) \emph{faster} – designing efficient decoding strategies to accelerate inference \revise{of reasoning models}. A curated collection of papers discussed in this survey is available in our GitHub repository: \url{https://github.com/fscdc/Awesome-Efficient-Reasoning-Models}.
\end{abstract}

\section{Introduction}
\label{sec:introduction}

Recent reasoning-oriented models, or Large Reasoning Models (LRMs)~\citep{guo2025deepseek,jaech2024openai}, have achieved remarkable performance on complex reasoning tasks by generating long Chain-of-Thoughts (CoTs), enabling effective problem-solving in domains such as mathematics and coding~\citep{sprague2024cot}. However, while LRMs significantly improve performance on reasoning tasks, they also cause substantial overhead. Compared to standard Large Language Models (LLMs), reasoning models lead to redundancy across multiple dimensions. 

\begin{figure*}[t]
\centering
  \includegraphics[width=0.96\linewidth, trim=10 10 10 10, clip]{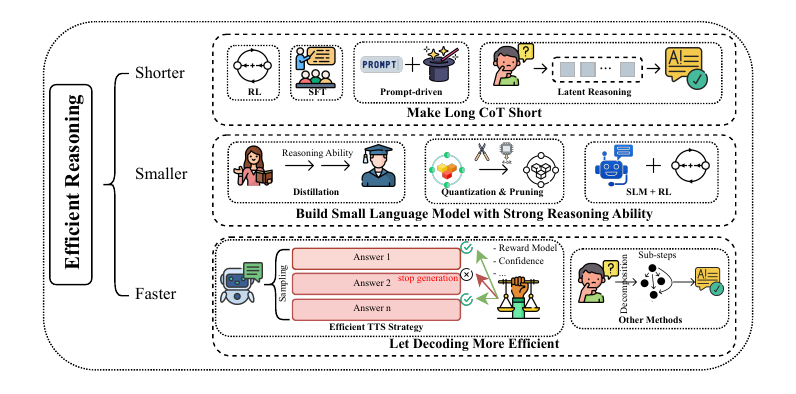} \\
\caption{Overview of efficient reasoning. We categorize existing efficient reasoning methods into three key directions based on how they improve reasoning efficiency: (1) make long CoT short (\emph{shorter}); (2) build small language models with strong reasoning ability (\emph{smaller}); and (3) let decoding more efficient (\emph{faster}).}
\label{fig:overview}
\end{figure*}

A salient characteristic of reasoning models is their tendency to overthink by generating excessively long reasoning chains~\citep{chen2024not,sui2025stop}, which has naturally motivated efforts to improve efficiency by shortening reasoning paths. Meanwhile, recent studies~\citep{wu2025more,yang2025towards,jin2024impact} challenge the assumption that longer CoTs always lead to better performance, showing even negative returns. To address this kind of CoT length redundancy, a range of methods have been proposed: reinforcement learning (RL) with length penalty~\citep{luo2025o1,aggarwal2025l1}, supervised fine-tuning (SFT) on variable-length CoT data~\citep{ma2025cot,xia2025tokenskip}, and prompt-driven strategies that either guide reasoning paths or route inputs to more efficient solutions~\citep{ding2024break,aytes2025sketch}. Furthermore, latent reasoning performs the process in latent space without generating explicit CoTs, making reasoning chains more concise~\citep{hao2024training,su2025token}.

In addition to excessively long reasoning chains, reasoning models typically rely on large model sizes to achieve strong reasoning performance (e.g., DeepSeek R1~\citep{guo2025deepseek} has 685B parameters), which leads to substantial computational and memory costs. To address this, model compression~\citep{han2015deep} has proven effective in reducing model size redundancy in standard LLMs, naturally inspiring interest in how these techniques (e.g., distillation~\citep{hinton2015distilling}, quantization~\citep{gray1998quantization}, and pruning~\citep{lecun1989optimal}) can be applied to improve reasoning efficiency. In parallel, another line of work directly builds small language models with strong reasoning abilities using  RL~\citep{li2023mixed,li2025small,zhu2024improving}.

Beyond length and model size redundancy, inefficiency can also arise during the decoding stage. A growing body of work focuses on accelerating inference through more efficient decoding strategies to tackle this issue. Test-time scaling (TTS) strategies, while enhancing reasoning performance~\citep{snell2024scaling}, also introduce latency redundancy during the decoding stage. Some methods~\citep{sun2024fast,wang2024make} specifically target and optimize the speed of certain TTS strategies~\citep{wang2022self}. Other approaches, like parallel decoding~\citep{ning2023skeleton} and problem decomposition~\citep{teng2025atom}, also mitigate inefficiency.

This survey aims to provide an overview of research in efficient reasoning. As illustrated in Figure~\ref{fig:overview}, we categorize existing works into three key directions based on the type of redundancy they target: (1) making long CoT short (\emph{shorter}), which focuses on enabling models to produce shorter reasoning paths while maintaining performance; (2) building small language model with strong reasoning abilities (\emph{smaller}), which aims to endow compact models with the ability to solve complex reasoning tasks; (3) making decoding more efficient (\emph{faster}), which explores strategies to reduce latency during the decoding stage.

The following sections of this survey cover the content as outlined below. Section~\ref{sec:background} will explore key backgrounds closely related to efficient reasoning. Section~\ref{sec:efficient-reasoning} will systematically introduce various methods and their relationships across three categories. Section~\ref{sec:metric-benchmark} presents the evaluation metrics, as well as datasets and benchmarks. Section~\ref{sec:challenges-future-direction} will discuss the key challenges in the field and propose some potential future research directions, while Section~\ref{sec:conclusion} will conclude the survey. Additionally, Figure~\ref{fig:taxonomy-efficient-reasoning} illustrates the taxonomy of efficient reasoning methods discussed in this survey.

\tikzstyle{my-box}=[
 rectangle,
 rounded corners,
 text opacity=1,
 minimum height=1.5em,
 minimum width=5em,
 inner sep=2pt,
 align=center,
 fill opacity=.8,
 draw
]
\tikzstyle{leaf}=[
 my-box,
 minimum height=1.5em,
 text=black,
 align=left,
 font=\scriptsize,
 inner xsep=2pt,
 inner ysep=4pt,
]
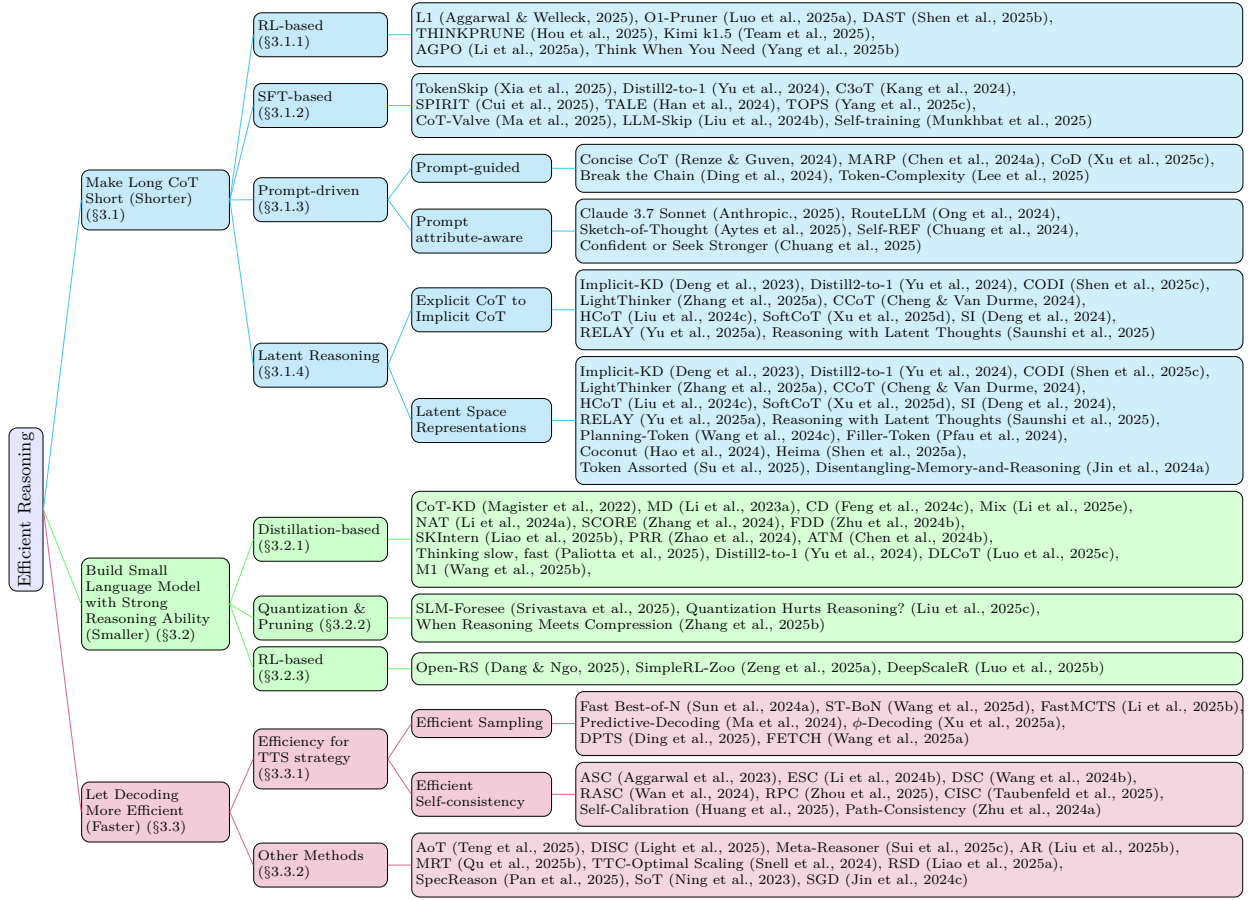
\begin{figure*}[t]
	\centering
	\resizebox{\textwidth}{!}{
		\begin{forest}
			for tree={
				grow=east,
				reversed=true,
				anchor=base west,
				parent anchor=east,
				child anchor=west,
				base=left,
				font=\small,
				rectangle,
				draw=black,
				rounded corners,
				align=left,
				minimum width=4em,
				s sep=3pt,
				inner xsep=2pt,
				inner ysep=3pt,
				ver/.style={rotate=90, child anchor=north, parent anchor=south, anchor=center},
			},
			where level=1{text width=7.0em,font=\scriptsize}{},
			where level=2{text width=6.3em,font=\scriptsize}{},
			where level=3{text width=6.6em,font=\scriptsize}{},
                where level=4{text width=6.6em,font=\scriptsize}{},
			[
			Efficient Reasoning, ver, fill=blue!10, edge={blue!40}
                [
			Make Long CoT \\ Short (Shorter) \\ ($\S$\ref{sec:cot-long2short})
                , fill=cyan!20, edge={cyan!60}
			[
			RL-based \\($\S$\ref{sec:rl-based-efficiency})
            , fill=cyan!20, edge={cyan!60}
			[
                L1~\citep{aggarwal2025l1}{,} 
                O1-Pruner~\citep{luo2025o1}{,}
                DAST~\citep{shen2025dast}{,} \\
                THINKPRUNE~\citep{hou2025thinkprune}{,}
                Kimi k1.5~\citep{team2025kimi}{,} \\
                AGPO~\citep{li2025adaptive}{,}
                Think When You Need~\citep{yang2025think} 
			, leaf, text width=41.2em, fill=cyan!20, edge={cyan!60}
			]
			]
			[
			SFT-based \\($\S$\ref{sec:sft-based-efficiency})
            , fill=cyan!20, edge={cyan!60}
			[
                TokenSkip~\citep{xia2025tokenskip}{,} 
                Distill2-to-1~\citep{yu2024distilling}{,} 
                C3oT~\citep{kang2024c3ot}{,} \\
                SPIRIT~\citep{cui2025stepwise}{,} 
                TALE~\citep{han2024token}{,} 
                TOPS~\citep{yang2025towards}{,} \\
                CoT-Valve~\citep{ma2025cot}{,} 
                LLM-Skip~\citep{liu2024can}{,}
                Self-training~\citep{munkhbat2025self}
			, leaf, text width=41.2em, fill=cyan!20, edge={cyan!60}
			]
			]
			[
			Prompt-driven \\($\S$\ref{sec:prompt-help-efficiency})
            , fill=cyan!20, edge={cyan!60}
			[
			Prompt-guided
            , fill=cyan!20, edge={cyan!60}
			[
                Concise CoT~\citep{renze2024benefits}{,} 
                MARP~\citep{chen2024unlocking}{,} 
                CoD~\citep{xu2025chain}{,} \\
                Break the Chain~\citep{ding2024break}{,} 
                Token-Complexity~\citep{lee2025well}
			, leaf, text width=33em, fill=cyan!20, edge={cyan!60}
			]
			]
			[
			Prompt \\ attribute-aware
            , fill=cyan!20, edge={cyan!60}
			[
                Claude 3.7 Sonnet~\citep{claude2025}{,} 
                RouteLLM~\citep{ong2406routellm}{,} \\
                Sketch-of-Thought~\citep{aytes2025sketch}{,} 
                Self-REF~\citep{chuanglearning}{,} \\
                Confident or Seek Stronger~\citep{chuang2025confident}
			, leaf, text width=33em, fill=cyan!20, edge={cyan!60}
			]
			]
			]
			[
			Latent Reasoning \\($\S$\ref{sec:latent-reasoning})
            , fill=cyan!20, edge={cyan!60}
                [
                Explicit CoT to \\ Implicit CoT 
                , fill=cyan!20, edge={cyan!60}
			[
                Implicit-KD~\citep{deng2023implicit}{,}
                Distill2-to-1~\citep{yu2024distilling}{,}
                CODI~\citep{shen2025codi}{,} \\
                LightThinker~\citep{zhang2025lightthinker}{,}
                CCoT~\citep{cheng2024compressed}{,} \\
                HCoT~\citep{liu2024expediting}{,} 
                SoftCoT~\citep{xu2025softcot}{,}
                SI~\citep{deng2024explicit}{,} \\
                RELAY~\citep{yu2025enhancing}{,} 
                Reasoning with Latent Thoughts~\citep{saunshi2025reasoning}
			, leaf, text width=33em, fill=cyan!20, edge={cyan!60}
			]
			]
			[
			Latent Space \\ Representations
            , fill=cyan!20, edge={cyan!60}
			[
                Implicit-KD~\citep{deng2023implicit}{,}
                Distill2-to-1~\citep{yu2024distilling}{,}
                CODI~\citep{shen2025codi}{,} \\
                LightThinker~\citep{zhang2025lightthinker}{,}
                CCoT~\citep{cheng2024compressed}{,} \\
                HCoT~\citep{liu2024expediting}{,} 
                SoftCoT~\citep{xu2025softcot}{,}
                SI~\citep{deng2024explicit}{,} \\
                RELAY~\citep{yu2025enhancing}{,} 
                Reasoning with Latent Thoughts~\citep{saunshi2025reasoning}{,} \\
                Planning-Token~\citep{wang2023guiding}{,}
                Filler-Token~\citep{pfau2024let}{,} \\
                Coconut~\citep{hao2024training}{,} 
                Heima~\citep{shen2025efficient}{,} \\
                Token Assorted~\citep{su2025token}{,}
                Disentangling-Memory-and-Reasoning~\citep{jin2024disentangling}
			, leaf, text width=33em, fill=cyan!20, edge={cyan!60}
			]
			]
			]
			]
			[
			Build Small \\ Language Model \\ with Strong \\ Reasoning Ability \\(Smaller) ($\S$\ref{sec:slm-with-reasoning-ability})
            , fill=green!20, edge={green!60}
			[
			Distillation-based 	\\($\S$\ref{sec:d-transfer-reasoning-ability})
            , fill=green!20, edge={green!60}
			[
                CoT-KD~\citep{magister2022teaching}{,} 
                MD~\citep{li2023mixed}{,} 
                CD~\citep{feng2024teaching}{,}
                Mix~\citep{li2025small}{,} \\
                NAT~\citep{li2024turning}{,} 
                SCORE~\citep{zhang2024small}{,} 
                FDD~\citep{zhu2024improving}{,} \\
                SKIntern~\citep{liao2025skintern}{,} 
                PRR~\citep{zhao2024probe}{,} 
                ATM~\citep{chen2024distilling}{,} \\
                Thinking slow{,} fast~\citep{paliotta2025thinking}{,}
                Distill2-to-1~\citep{yu2024distilling}{,}
                DLCoT~\citep{luo2025deconstructing}{,} \\
                M1~\citep{wang2025m1}{,} 
			, leaf, text width=41.2em, fill=green!20, edge={green!60}
			]
			]
			[
			Quantization \& \\ Pruning ($\S$\ref{sec:pruning-quant-retain-reasoning-ability})
            , fill=green!20, edge={green!60}
			[
                SLM-Foresee~\citep{srivastava2025towards}{,}
                Quantization Hurts Reasoning{?}~\citep{liu2025quantization}{,} \\
                When Reasoning Meets Compression~\citep{zhang2025reasoning}
			, leaf, text width=41.2em, fill=green!20, edge={green!60}
			]
			]
			[
			RL-based \\($\S$\ref{sec:build-slm-with-rl})
            , fill=green!20, edge={green!60}
			[
                Open-RS~\citep{dang2025reinforcement}{,}
                SimpleRL-Zoo~\citep{zeng2025simplerl}{,}
                DeepScaleR~\citep{luo2025deepscaler}
			, leaf, text width=41.2em, fill=green!20, edge={green!60}
			]
			]
			]
			[
			Let Decoding \\ More Efficient \\ (Faster) ($\S$\ref{sec:decoding-tricks})
            , fill=purple!20, edge={purple!60}
			[
			Efficiency for \\ TTS strategy 	\\($\S$\ref{sec:tts-efficiency})
            , fill=purple!20, edge={purple!60}
                [
			Efficient Sampling
            , fill=purple!20, edge={purple!60}
			[
                Fast Best-of-N~\citep{sun2024fast}{,}
                ST-BoN~\citep{wang2025sampling}{,}
                FastMCTS~\citep{li2025fastmcts}{,} \\
                Predictive-Decoding~\citep{ma2024non}{,}
                $\phi$-Decoding~\citep{xu2025phi}{,} \\
                DPTS~\citep{ding2025dynamic}{,}
                FETCH~\citep{wang2025don}
			, leaf, text width=33em, fill=purple!20, edge={purple!60}
			]
			]
			[
			Efficient \\Self-consistency
                , fill=purple!20, edge={purple!60}
			[
                ASC~\citep{aggarwal2023let}{,}
                ESC~\citep{li2024escape}{,}
                DSC~\citep{wang2024make}{,} \\
                RASC~\citep{wan2024reasoning}{,}
                RPC~\citep{zhou2025bridging}{,}
                CISC~\citep{taubenfeld2025confidence}{,} \\
                Self-Calibration~\citep{huang2025efficient}{,}
                Path-Consistency~\citep{zhu2024path}
			, leaf, text width=33em, fill=purple!20, edge={purple!60}
			]
			]
			]
			[
			Other Methods \\($\S$\ref{sec:optimal-ttc})
                , fill=purple!20, edge={purple!60}
			[
			AoT~\citep{teng2025atom}{,}
                DISC~\citep{light2025disc}{,}
                Meta-Reasoner~\citep{sui2025meta}{,}
                AR~\citep{liu2025chaos}{,} \\
                MRT~\citep{qu2025optimizing}{,}
                TTC-Optimal Scaling~\citep{snell2024scaling}{,}
                RSD~\citep{liao2025reward}{,} \\
                SpecReason~\citep{pan2025specreason}{,}
                SoT~\citep{ning2023skeleton}{,}
                SGD~\citep{jin2024adaptive}
			, leaf, text width=41.2em, fill=purple!20, edge={purple!60}
			]
			]
                ]
			]
		\end{forest}
	}
	\caption{Taxonomy of efficient reasoning.}
	\label{fig:taxonomy-efficient-reasoning}
\end{figure*}

\section{Background}
\label{sec:background}

\subsection{Chain-of-Thought Reasoning}
\label{sec:cot-reasoning}

CoT~\citep{wei2022chain} serves as a baseline reasoning approach, enabling LLMs to generate a sequence of intermediate steps before reaching the final answer, thus significantly improving performance on complex reasoning tasks. Various extensions have subsequently been proposed to further enhance reasoning capabilities. For instance, Tree-of-Thought (ToT)~\citep{yao2023tree} generalizes the linear CoT structure into a tree, facilitating the exploration of multiple reasoning paths through backtracking and lookahead strategies. Graph-of-Thoughts (GoT)~\citep{besta2024graph} has expanded this approach into graph structures to better capture dependencies and compositional relationships among reasoning steps, substantially improving reasoning quality. Additionally, some specialized CoT variants are task-specific. PoT~\citep{chen2022program} disentangles reasoning from computation by having the language model generate programmatic reasoning steps (i.e., expressing thoughts as code), which an external calculator executes to obtain the final answer, making this approach particularly effective for math and financial tasks. CoS~\citep{hu2024chain}, on the other hand, targets spatial reasoning by leveraging compressed symbolic representations of spatial relations to reduce token usage.

\subsection{\revise{Reasoning Models and Underlying Techniques}}
\label{sec:tech-lrm}

\revise{Recent reasoning models have moved beyond early prompting-based CoT techniques by internalizing step-by-step reasoning through SFT and RL. Building structured reasoning paradigms mentioned in Section~\ref{sec:cot-reasoning}, these models are trained to generate reasoning traces aligned with human-like logic. RL plays a crucial role by optimizing for reasoning quality using reward signals based on correctness, format alignment, and process supervision~\citep{xu2025towards,ouyang2022training,zhou2022least}. Advanced models like OpenAI o1~\citep{openaio1} are believed to incorporate tree-search strategies~\citep{coulom2006efficient} and process reward models to guide the exploration of intermediate steps. Others, such as DeepSeek R1~\citep{guo2025deepseek}, employ rule-based reward functions to reinforce correct reasoning steps.}

\subsection{\revise{Test-Time Scaling}}
\label{sec:tts}

Scaling test-time computation (TTC) is another road for enhancing reasoning performance~\citep{snell2024scaling,zeng2025revisiting}. \revise{Scaling can be approached from two complementary dimensions: horizontal and vertical. The horizontal perspective involves generating multiple samples and selecting the best answer. Best-of-N~\citep{cobbe2021training,sun2024fast} selects the top-scoring response, while self-consistency~\citep{wang2022self} identifies the most consistent answer across reasoning chains. The vertical perspective focuses on increasing the length of a single reasoning path. For example, Self-Refine~\citep{madaan2023self} iteratively improves an initial response via self-evaluation, while other works~\citep{chen2023teaching,gou2023critic} leverage external feedback to guide the refinement process.} Additionally, an empirical study~\citep{wu2024inference} investigates the trade-offs between the efficiency and performance of various TTS strategies (e.g., Best-of-N, weighted voting) under different model sizes and computation budgets, providing practical insights for further research and deployment.

\subsection{\revise{Model Compression}}
\label{sec:model-compression}

\revise{Model compression strategies are widely used to reduce the size and computational overhead of models~\citep{han2015deep}. Common approaches include quantization~\citep{gray1998quantization,frantar-gptq,lin2023awq,xiao2023smoothquant}, which reduces model size by lowering the precision of model parameters. Pruning~\citep{lecun1989optimal,ma2023llmpruner,fang2023depgraph,wang2020neural} removes less significant or redundant model parameters to achieve sparsity, reducing model size and inference latency. Unlike the above techniques, knowledge distillation~\citep{hinton2015distilling,wang2022selfinstruct,liu2019knowledge} achieves compression not by directly modifying the original model, but by transferring knowledge from a larger, well-trained teacher model to a smaller student model, allowing the student to replicate the teacher’s behavior while maintaining comparable performance (see details about model compression in Appendix~\ref{apx:model-compression-details}).}

\begin{figure*}[t]
\centering
  \includegraphics[width=0.96\linewidth, trim=10 5 10 5, clip]{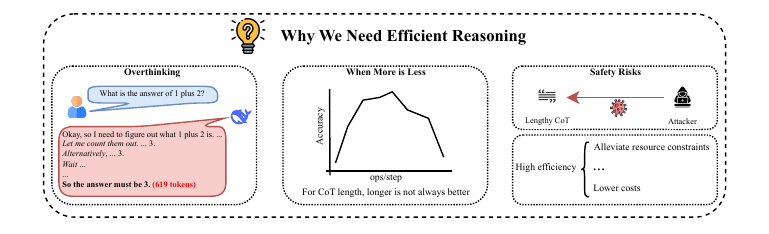} \\
\caption{Motivation for efficient reasoning. \textit{(Left)} Models often exhibit overthinking, generating unnecessarily long reasoning chains even for simple tasks. \textit{(Middle)} Longer reasoning is not always better and may result in reduced accuracy when excessively verbose. \textit{(Right)} Lengthy reasoning increases computational costs and poses safety risks. In addition, improving efficiency helps alleviate resource constraints and lower costs.}
\label{fig:motivation}
\end{figure*}

\subsection{Why We Need Efficient Reasoning}
\label{sec:why-efficient-reasoning}

Efficiency is a valuable research direction across many fields, and in the context of reasoning, we highlight key motivations for pursuing efficient reasoning (see Figure~\ref{fig:motivation}). Reasoning models often generate excessively long reasoning chains to solve reasoning tasks, even for simple samples, and typically rely on larger model sizes to achieve stronger reasoning performance. \revise{For example, answering ``What is the answer of 1 plus 2?'' requires 619 tokens from DeepSeek R1-685B (see Appendix~\ref{apx:overthinking-example} for details). To further illustrate the overhead, we evaluated four versions of DeepSeek R1 on the AIME~24 dataset and observed consistently huge token counts: 15513 for 1.5B, 12377 for 7B, 10854 for 14B, and 10024 for 32B.} Additionally, some strategies, such as Best-of-N and self-consistency, further scale the decoding process to enhance reasoning performance. These lead to substantial computational and memory demands. Moreover, overly long reasoning paths can accumulate errors and negatively impact final accuracy~\citep{wu2025more,yang2025towards}.

On the other hand, efficient reasoning is also essential in real-world applications such as embodied AI~\citep{duan2022survey}, agent systems~\citep{wang2024survey}, and real-time platforms (e.g., autonomous driving~\citep{cui2024survey}). In these scenarios, efficiency enables agents to process sensory inputs in real time, make swift and accurate decisions, and interact seamlessly with dynamic environments. Additionally, unnecessarily lengthy reasoning may increase safety risks~\citep{kuo2025h,li2025output}, posing unpredictable threats. These challenges collectively highlight the limitations of current reasoning models, underscoring the necessity of improving reasoning efficiency.

\begin{table*}[!t]
\centering
\caption{Performance of efficient reasoning methods on the AIME~24 dataset. $\dagger$ denotes the result of the original model, averaged over 5 independent runs.}
\label{tab:aime-result}
\vspace{2mm}
\resizebox{0.995\linewidth}{!}{
\setlength{\tabcolsep}{5mm}
\begin{tabular}{lllll} 
\toprule
\bf Category & \bf Type & \bf Methods & \bf Acc. / \#Tokens & \bf Base Model \\
\midrule
Original Model & - & Baseline$^\dagger$ & 70.67\% / 10024 & DeepSeek-R1-32B \\
\hdashline
Shorter & RL  & DAST & 53.30\% / 6337 & DeepSeek-R1-Distill-Qwen-7B \\
Shorter & SFT  & CoT-Valve & 43.30\% / 4630 & QwQ-32B-Preview \\
Shorter & SFT  & TOPS & 46.00\% / 6427 & Qwen2.5-32B \\
\hdashline
Smaller & KD & Mix & 10.00\% / - & Qwen2.5-3B \\
Smaller & KD & DLCoT & 53.30\% / 18825 & Qwen2.5-14B \\
Smaller & RL & Open-RS  & 46.70\% / - & DeepSeek-R1-Distill-Qwen-1.5B \\
Smaller & RL & DeepSacleR & 43.10\% / - & DeepSeek-R1-Distill-Qwen-1.5B \\ 
\hdashline
Faster & Efficient self-consistency & RPC & 9.50\% / - & InternLM-2-MATH-Plus 7B \\
Faster & Efficient sampling & $\phi$-Decoding & 16.67\% / - & LLaMA3.1-8B-I \\
\bottomrule
\end{tabular}}
\end{table*}

\section{Efficient Reasoning}
\label{sec:efficient-reasoning}

In the following, we introduce efficient reasoning methods based on three key categories: shortening long chains of thought, as discussed in Section~\ref{sec:cot-long2short}; developing small language models with strong reasoning capabilities, details of which can be found in Section~\ref{sec:slm-with-reasoning-ability}; and improving decoding efficiency, which is elaborated in Section~\ref{sec:decoding-tricks}. We present the performance of various efficient reasoning methods on the challenging AIME~24 dataset in Table~\ref{tab:aime-result} \revise{and further provide a latency-based summary of representative methods across categories on the GSM8K dataset in Table~\ref{tab:gsm8k-all-category}.}

\subsection{Make Long CoT Short}
\label{sec:cot-long2short}

\revise{Recent works have explored various approaches to improve reasoning efficiency by shortening CoT length without compromising reasoning performance.} Among them, RL with length penalty is widely used for encouraging concise and effective reasoning paths (see Section~\ref{sec:rl-based-efficiency}). Another line of work explores SFT with variable-length CoT data to improve reasoning efficiency, as discussed in Section~\ref{sec:sft-based-efficiency}. In addition, prompt-driven techniques improve reasoning efficiency by utilizing prompts, with further details available in Section~\ref{sec:prompt-help-efficiency}. Finally, we explore latent reasoning, which performs the reasoning process in latent space and drastically reduces CoT length, with details provided in Section~\ref{sec:latent-reasoning}. \revise{Additionally, Table~\ref{tab:make-long-cot-short} provides an overview of these methods, showing that most RL-based methods utilize Full FT, while many SFT-based methods adopt Parameter-Efficient Fine-Tuning (PEFT) techniques like LoRA~\citep{hu2022lora} to reduce cost. This trend suggests that RL-based methods require more extensive parameter updates, making lightweight adaptation less effective; for latent reasoning, Full FT remains dominant, and these methods often yield higher speedups, indicating that implicit representations enable more effective compression and offer a higher upper bound compared to explicit reasoning chains.}

\subsubsection{Reinforcement Learning Helps Efficiency Improvement}
\label{sec:rl-based-efficiency}

Incorporating explicit chain length penalty into RL is a natural strategy for shortening reasoning chains~\citep{team2025kimi,li2025adaptive,arora2025training}. L1~\citep{aggarwal2025l1} takes this further by introducing designated length-constraint instructions into the training data. O1-Pruner~\citep{luo2025o1} develops a specialized reward design by utilizing length and accuracy from a reference model as baselines, explicitly rewarding shorter reasoning paths and higher accuracy to ensure efficiency without sacrificing performance. DAST~\citep{shen2025dast} aims to achieve a balanced CoT (i.e., dynamically adjusting computational resources by allocating more reasoning steps to more challenging questions and fewer to simpler ones). Specifically, it proposes a Token Length Budget (TLB), defined as a weighted sum of the mean token count in accurate answers and a predefined upper bound on generation length to quantify problem difficulty, penalizing excessively verbose reasoning for simple questions while encouraging comprehensive reasoning for complex ones. THINKPRUNE~\citep{hou2025thinkprune} designs a length-aware reward function that only provides a reward if the correct answer is generated within a specified token budget. The model is trained using the Group Relative Policy Optimization (GRPO) algorithm with progressively tightened length constraints. Additionally, Think When You Need~\citep{yang2025think} utilizes pairwise comparisons to generate rewards based on the relative length and accuracy of reasoning, guiding models to produce concise yet accurate solutions.

\subsubsection{Supervised Fine-Tuning with Variable-Length CoT Data Helps Efficiency Improvement}
\label{sec:sft-based-efficiency}

\begin{table*}[!t]
\centering
\caption{Overview of efficient reasoning methods in Section~\ref{sec:cot-long2short}. The speedup ratio is computed by comparing either the latency (L.) or the token count (T.). $Avg_1$ represents the average of Llama-3.2-3B, Gemma2-2B, Qwen2.5-3B, Qwen2.5-Math-1.5B, and DeepSeekMath-7B; $Avg_2$ represents the average of GPT-4o, GPT-4o-mini, Yi-lightning, o3-mini, and LLaMA3.1-8B-I.}
\label{tab:make-long-cot-short}
\vspace{2mm}
\resizebox{0.995\linewidth}{!}{
\setlength{\tabcolsep}{0.5mm}
\begin{tabular}{lllllr}
\toprule
\bf Type & \bf Methods & \bf Training Scheme & \bf Acc. / \#Tokens & \bf Base Model & \bf Speedup \\
\midrule
RL  & O1-Pruner & PPO (Freeze FT) & GSM8K: 96.50\% / 543 & QwQ-32B & 1.5 - 2.0 $\times$ (L.)  \\ 
RL  & DAST & SimPO (Full FT) & MATH-500: 92.60\% / 2802  & DeepSeek-R1-Distill-Qwen-7B & 1.6 - 2.2 $\times$ (T.) \\ 
RL  & AGPO & GRPO (Full FT) & MATH-500: 77.20\% / 463  & Qwen2.5-Math-7B & 1.3 - 1.5 $\times$ (T.)  \\ 
RL  & THINKPRUNE & GRPO (Full FT) & MATH-500: 83.90\% / 2209  & DeepSeek-R1-Distill-Qwen-1.5B & 1.7 - 2.0 $\times$ (T.)  \\ 
RL  & Think When You Need & GRPO (Full FT) & -  & - & 1.3 $\times$ (T.)  \\ 
\hdashline
SFT  & TokenSkip &  SFT (LoRA) & GSM8K: 78.20\% / 113  & LLaMA3.1-8B-I  & 1.7 - 1.8 $\times$ (L.)   \\ 
SFT  & C3oT &  SFT (Full FT) & GSM8K: 47.10\% / -  & LLaMA2-Chat-13B & 2.0 $\times$ (T.)  \\ 
SFT  & Self-Training &  SFT (Full FT) & GSM8K: 78.07\% / 176  & $Avg_1$ & 1.3 - 1.5 $\times$ (T.) \\ 
SFT  & TALE &  SFT / DPO (LoRA) & GSM8K: 78.57\% / 140  & $Avg_2$ & 1.7 $\times$ (T.) \\ 
SFT  & CoT-Valve & Progressive SFT (LoRA) & GSM8K: 95.40\% / 289  & QwQ-32B & 2.6 $\times$ (T.) \\ 
\hdashline
Prompting  & Concise CoT & Training-free & - & - &  1.9 - 2.0 $\times$ (T.) \\ 
Prompting  & Break the Chain & Training-free & GSM8K: 74.22\% / -  & ChatGPT & - \\ 
Prompting  & TALE-EP & Training-free & GSM8K: 84.46\% / 77 &  GPT-4o-mini &  4.1 $\times$ (T.) \\ 
Prompting  & CoD & Training-free & GSM8K: 91.10\% / 44  & GPT-4o & 4.7 $\times$ (T.)  \\ 
\hdashline
Routing  & RouteLLM & LLaMA3-8B Router & GSM8K: 74.82\% / - &  GPT-4  &  1.5 $\times$ (T.) \\ 
Routing & Sketch-of-Thought & DistillBERT Router & - & - & 3.6 $\times$ (T.) \\
Routing & Self-REF &  SFT (LoRA) & GSM8K: 81.60\% / -  & LLaMA3-8B-I & 1.2 - 2.0 $\times$ (L.)  \\
\hdashline
Latent reasoning & Implicit-KD &  SFT (Full FT)& GSM8K: 20.00\% / -  & GPT-2 small & 8.2 $\times$ (L.)  \\
Latent reasoning & SI & Progressive SFT (Full FT)& GSM8K: 30.00\% / -  & GPT-2 small & 4.0 - 11.0 $\times$ (L.)  \\
Latent reasoning & CCoT &  SFT (LoRA) & GSM8K: 17.90\% / -  & CCOT 
\& DECODE & 10.4 - 24.5 $\times$ (L.)  \\
Latent reasoning & SoftCoT &  SFT (Freeze FT) & GSM8K: 85.81\% / -  & Qwen2.5-7B-I & 4.0 - 5.0 $\times$ (L.)  \\
Latent reasoning & CODI & Self-distillation (LoRA)& GSM8K: 43.70\% / -  & GPT-2 small & 2.5 - 2.7 $\times$ (L.)  \\
Latent reasoning & LightThinker &  SFT (Full FT)& GSM8K: 90.14\% / -  & Qwen2.5-7B & up to 1.4 $\times$ (L.) \\
Latent reasoning & Coconut & Progressive SFT (Full FT)& GSM8K: 34.10\% / 8  & GPT-2 & 3.0 $\times$ (T.) \\
Latent reasoning & Token Assorted & SFT (Full FT) & GSM8K: 84.10\% / 194  & LLaMA3.1-8B & 1.2 $\times$ (T.) \\
\bottomrule
\end{tabular}}
\end{table*}

Following a clear fine-tuning pipeline, we organize the discussion of this line of research into two stages: (1) how variable-length CoT data is constructed and (2) which SFT approach (i.e., standard or progressive) is adopted. For each work, we explicitly address these two questions to facilitate comparison and analysis.

\paragraph{How variable-length CoT data is constructed?} 
To construct variable-length CoT data, long reasoning chains are commonly generated by prompting LLMs with inputs, whereas the key challenge lies in obtaining the corresponding shorter reasoning chains. To address this, existing approaches generally fall into two categories. The first approach involves compressing existing long reasoning paths into shorter ones. For instance, TokenSkip~\citep{xia2025tokenskip} identifies and skips less important tokens based on their semantic contribution to the final answer. Distill2-to-1~\citep{yu2024distilling} discards reasoning steps entirely, retaining only high-quality (input, answer) pairs through consistency-based filtering. C3oT~\citep{kang2024c3ot} leverages GPT-4 as a compressor to shorten chain length by preserving essential reasoning details. Additionally, SPIRIT~\citep{cui2025stepwise} uses perplexity to evaluate step importance, thus selectively compressing reasoning paths. 

The alternative approach directly generates short reasoning paths. Self-training~\citep{munkhbat2025self} employs multiple sampling combined with few-shot prompting, selecting the shortest correct reasoning paths. TALE~\citep{han2024token} observes that LLMs naturally follow token budget constraints specified in prompts and introduces a binary search-based algorithm to identify the optimal token budget for generating concise reasoning paths. TOPS~\citep{yang2025towards} begins with a small set of o1-like responses (i.e., either generated by existing models or manually constructed) as seed data. Each response corresponds to a different level of reasoning effort. Using this data, it trains a tag model that learns to produce variable-length reasoning paths conditioned on effort-specific prompts, enabling the construction of diverse CoT data with controllable lengths. Inspired by model merging~\citep{yang2024model}, CoT-Valve~\citep{ma2025cot} achieves chain length control by adjusting a specific direction of the parameter space, merging parameters from a base LLM with those of a reasoning-enhanced model of identical architecture\footnote{Model merging is an effective strategy for efficient reasoning. For example, Kimi k1.5~\citep{team2025kimi} improves token efficiency by merging a long-cot model and a short-cot model, while \citet{wu2025unlocking} combines System 1 and System 2 models to shorten response length.}. Additionally, LLM-Skip~\citep{liu2024can} manually shortens reasoning paths for complex datasets at the initial training stage, explicitly labeling prompts with ``Solve it in n steps.''. In the subsequent progressive SFT process, shorter reasoning paths generated by the model are continuously integrated into the training set.

\paragraph{Which SFT approach is adopted?} 
Most works adopt a standard SFT approach~\citep{xia2025tokenskip,yu2024distilling,kang2024c3ot,cui2025stepwise,munkhbat2025self,han2024token,ma2025cot,yang2025towards}, typically leveraging either LoRA~\citep{xia2025tokenskip,ma2025cot} or full fine-tuning~\citep{kang2024c3ot}. Notably, C3oT~\citep{kang2024c3ot} designs a conditioned training strategy, enabling the model to learn both long and short reasoning styles during training and generate concise reasoning paths at inference by simply appending a short condition in the prompt. TALE~\citep{han2024token} further explores DPO as an alternative fine-tuning objective, allowing direct control over the model’s output preference.

Another line of work adopts progressive fine-tuning strategies~\citep{liu2024can,ma2025cot}. LLM-Skip~\citep{liu2024can} iteratively encourages the model to generate shorter reasoning paths and then merges the generated shorter paths into the training set for subsequent fine-tuning rounds, gradually reducing chain length. CoT-Valve~\citep{ma2025cot} supports both standard SFT and two progressive strategies: CoT-Valve++ and CoT-Valve+P. CoT-Valve++ introduces a normalized path-length factor $\beta$, which is smaller for longer paths. During training, the model parameters are dynamically adjusted along a direction scaled by $\beta$, allowing the model to adapt to reasoning paths of varying lengths and learn finer-grained length control. CoT-Valve+P, on the other hand, progressively trains the model on samples sorted from long to short chains, guiding it to shorten the chain length over successive fine-tuning stages.

\subsubsection{Prompt-Driven Efficiency Enhancement in Reasoning}
\label{sec:prompt-help-efficiency}

\revise{We categorize prompt-driven works into two directions}: (1) prompt-guided reasoning, which leverages well-designed prompts to guide reasoning models toward more effective reasoning paths and (2) prompt-based routing, which utilizes prompt-level attributes (e.g., complexity) to adaptively select appropriate computational paths (e.g., route easy questions to lightweight models and hard ones to powerful large models).

\paragraph{Prompt-guided Efficient Reasoning.}
\label{par:prompt-guided-efficiency}

Concise CoT~\citep{renze2024benefits} shows that simply adding ``Be concise'' to the prompt can shorten reasoning chains. Break the Chain~\citep{ding2024break} leverages carefully crafted instructions (e.g., ``rapidly evaluate and use the most effective reasoning shortcut'') to trigger the model’s ability to exploit shortcuts and skip unnecessary steps. TALE-EP~\citep{han2024token} employs an LLM-based estimator to predict the minimal token budget required for each question, which is then incorporated into the prompt to guide efficient reasoning. CoD~\citep{xu2025chain} develops the instruction ``Think step by step, but only keep a minimum draft for each thinking step, with 5 words at most.'', which significantly reduces token usage under few-shot settings without compromising accuracy. However, its performance degrades in zero-shot settings and on small language models. MARP~\citep{chen2024unlocking} boosts per-step information density and reduces step count under a fixed reasoning boundary, achieving high efficiency gains through prompt design, and can be further combined with PoT for better computation-reasoning separation. Token-Complexity~\citep{lee2025well} presents token complexity to measure the minimal tokens needed for correct reasoning and derives the theoretical compression limit of CoT chains. Through prompt variations (e.g., ``use 10 words or less'' or ``remove all punctuation''), they explore the trade-off between performance and efficiency and show that current methods still fall far from the optimal bound, leaving room for improvement. Additionally, these methods can effectively construct variable-length CoT data, thereby supporting the approaches introduced in Section~\ref{sec:sft-based-efficiency}.

\paragraph{Prompt Attribute-Aware Efficient Reasoning.}
\label{par:prompt-attribute-aware-efficiency}

Claude 3.7 Sonnet~\citep{claude2025} offers two response modes (e.g., quick answers or step-by-step thinking), allocating more compute to complex reasoning tasks. Although the implementation details remain undisclosed, it is the first hybrid reasoning model and a foundation for subsequent methods.

Routing strategies primarily fall into two categories: classifier-based and uncertainty-based. Classifier-based approaches train a separate router to categorize incoming questions and route them to the most suitable model. RouteLLM~\citep{ong2406routellm} trains a router using preference data to dispatch easy questions to lightweight and harder ones to stronger models. Sketch-of-Thought~\citep{aytes2025sketch} routes each input to the most appropriate reasoning pattern by referencing cognitive science~\citep{goel1995sketches}, introducing three heuristic modes: Conceptual Chaining, which links ideas using minimal language; Chunked Symbolism, which organizes reasoning into symbolic blocks; and Expert Lexicons, which leverage domain-specific shorthand.

Uncertainty-based methods rely on confidence to guide routing. Self-REF~\citep{chuanglearning} adds two special tokens (i.e., <CN> for confident and <UN> for unconfident) to indicate confidence, training the model on annotated responses to self-assess its confidence level. If uncertain, the model defers to a more potent model or abstains. Confident or Seek Stronger~\citep{chuang2025confident} further analyzes uncertainty-based routing, observing that uncertainty distributions are relatively stable across tasks but vary significantly across models and uncertainty quantification (UQ) methods. It further designs a calibrated data construction strategy that improves the reliability of routing decisions for small language models.

\subsubsection{Reasoning in Latent Space}
\label{sec:latent-reasoning}

\revise{Unlike explicit CoT reasoning, latent reasoning~\citep{deng2023implicit,tan2025think} performs the reasoning process in latent space, skipping the generation of explicit intermediate steps.} Latent reasoning brings two key benefits: it allows for more human-like thinking by modeling complex ideas beyond language, and improves efficiency by reducing the need for explicit reasoning chains. This section first examines how models transition from explicit to implicit reasoning. Then, we explore how reasoning is represented in latent space.

\paragraph{From Explicit CoT to Implicit CoT.}
\label{sec:explicit-cot-2-implicit-cot}

As the seminal work introducing implicit CoT, Implicit-KD~\citep{deng2023implicit} proposed a distillation-based framework where a student model learns to reason implicitly by mimicking the hidden states across different layers of an explicit CoT teacher. To eliminate the reliance on the teacher model during inference, they further trained a simulator that directly maps input to teacher hidden states. SI~\citep{deng2024explicit} progressively removes intermediate reasoning steps through SFT, enabling the model to internalize reasoning without explicit chains. Similarly, Distill2-to-1~\citep{yu2024distilling} showed that SFT on (input, answer) pairs alone can yield strong implicit reasoning capabilities. CODI~\citep{shen2025codi} introduces a novel self-distillation framework where a shared model acts both as teacher and student—explicit CoT is learned via language modeling, while implicit CoT is learned by aligning the hidden activation of the token intermediately preceding the answer. LightThinker~\citep{zhang2025lightthinker} proposes a dynamic compression strategy for CoT. It segments the reasoning chain and compresses each step into special tokens, with a focus on the KV cache compression. These latent representations are used for subsequent reasoning, with attention masks designed to ensure the model can only access compressed content rather than whole previous steps.

Another line of work explores using an auxiliary model to generate latent reasoning tokens directly from the input. CCoT~\citep{cheng2024compressed} trains a lightweight CCOT module (a LoRA~\citep{hu2022lora}) to produce compressed latent reasoning tokens directly from input, which are then fed into a decoding module to generate concise answers, while HCoT~\citep{liu2024expediting} adopts a similar pipeline but places greater emphasis on semantic alignment during compression. SoftCoT~\citep{xu2025softcot} adopts a similar strategy by training a lightweight assistant model to produce implicit representations conditioned on the input. Furthermore, Reasoning with Latent Thoughts~\citep{saunshi2025reasoning} demonstrated that looping a transformer multiple times could emulate a deeper model and naturally induce latent thoughts, effectively capturing iterative reasoning without tokenized steps. RELAY~\citep{yu2025enhancing} follows this idea by aligning each iteration of a looped transformer~\citep{giannou2023looped} with explicit CoT steps. The trained looped model is then leveraged to produce high-quality CoT chains to train stronger autoregressive models on long reasoning tasks.

\paragraph{Latent Space Representations for Reasoning.}
\label{sec:latent-representation}

A common choice for latent space representation is to use continuous tokens~\citep{zhang2025lightthinker,shen2025codi,cheng2024compressed,xu2025softcot,hao2024training,liu2024expediting}, which naturally align with the internal computation of neural networks. Coconut~\citep{hao2024training} models reasoning in the hidden space by feeding the final-layer hidden states back into the model without decoding explicit CoT tokens, enabling more continuous and efficient reasoning. This approach unlocks advantages that explicit CoT cannot offer, such as backtracking and parallel decoding. Inspired by Coconut, Heima~\citep{shen2025efficient} introduces thinking tokens into multimodal large language models (MLLMs) to replace explicit reasoning steps, enabling reasoning in the latent space.

Another alternative approach is to employ discrete tokens as explicit representations of intermediate reasoning stages. Planning-Token~\citep{wang2023guiding} employs a set of planning tokens inserted before each reasoning step to guide the model to generate a latent plan before producing the detailed explanation. These tokens are obtained by clustering the hidden states of reasoning steps, yielding semantically meaningful and distinct discrete representations. Filler-Token~\citep{pfau2024let} proposes inserting meaningless filler tokens (e.g., repeated dots) into the reasoning path, allowing the model to perform additional hidden computation, thereby enhancing performance on reasoning tasks. Token Assorted~\citep{su2025token} improves reasoning efficiency by mixing text tokens with latent tokens obtained through VQ-VAE~\citep{van2017neural}, reducing sequence length while preserving key information. Disentangling-Memory-and-Reasoning~\citep{jin2024disentangling} introduces explicit discrete markers such as ⟨memory⟩ and ⟨reason⟩, which enable the model to disentangle reasoning into separate phases (i.e., retrieving relevant knowledge and performing logical inference) within the latent space. This separation facilitates more structured and interpretable reasoning behaviors.

\subsection{Build Small Language Model with Strong Reasoning Ability}
\label{sec:slm-with-reasoning-ability}

Compared to compressing reasoning chains, an alternative approach to improving reasoning efficiency is to empower small language models (SLMs) with strong reasoning capabilities. Due to their lower memory and computational requirements, SLMs are inherently more efficient and easier to deploy in real-world applications. Model compression~\citep{han2015deep,frantar2022gptq,li2023sparse} naturally aligns with this goal, as it enables small or compressed models to retain or gain reasoning abilities. A natural starting point is to transfer reasoning capabilities from larger models via distillation (see Section~\ref{sec:d-transfer-reasoning-ability}). We further explore other model compression techniques, including pruning and quantization, which aim to compress models without severely compromising reasoning performance in Section~\ref{sec:pruning-quant-retain-reasoning-ability}. Beyond traditional model compression techniques, RL offers another promising direction, enhancing reasoning capabilities under limited resources through carefully designed training strategies, as discussed in Section~\ref{sec:build-slm-with-rl}. \revise{Additionally, a summary of these methods is presented in Table~\ref{tab:building-slm}, indicating that most distillation approaches still rely on Full FT, with a few adopting PEFT techniques. Notably, methods that progressively incorporate refined or synthesized data (e.g., FDD and SKIntern) tend to achieve greater performance improvements.}

Apart from model compression and RL, some studies explore the reasoning ability of small language models from alternative perspectives. For example, \citet{liu2025can} shows that small language models can match or even surpass the reasoning performance of much larger LLMs with carefully designed TTS strategies. \revise{However, the effectiveness of TTS strategies varies with model architecture, reward design, and task complexity. While small language models show potential in reasoning, their limitations in instruction following and self-reflection highlight the need for further adaptation to align with human intent.}


\begin{table*}[!t]
\centering
\caption{Overview of efficient reasoning methods in Section~\ref{sec:slm-with-reasoning-ability}. $\text{Blended}_1$ represents the combination of s1 and DeepSacleR datasets; $\text{Blended}_2$ represents the combination of Omni-MATH, AIME, AMC, and Still datasets.}
\label{tab:building-slm}
\vspace{2mm}
\resizebox{0.995\linewidth}{!}{
\setlength{\tabcolsep}{0.5mm}
\begin{tabular}{llllll} 
\toprule
\bf Type & \bf Methods & \bf Training Scheme & \bf Training Data & \bf Acc. & \bf Base Model \\
\midrule
KD  & CoT-KD & Distillation (Full FT) &  CoT data & GSM8K: 21.99\% ($\uparrow$~13.88\%) & T5 XXL  \\ 
KD & MD & Mixed distillation (Freeze FT) & CoT and PoT data & GSM8K: 41.50\% ($\uparrow$~28.20\%) & LLaMA2-7B \\
KD & Mix & Mixed distillation (Full FT \& LoRA) & Long and short CoT data & GSM8K: 79.20\% ($\uparrow$~1.70\%) & LLaMA3.2-3B \\
KD & NAT & Mixed distillation (LoRA) & Positive and negative data & GSM8K: 41.24\% ($\uparrow$~23.73\%) & LLaMA-7B \\
KD & CD & Counterfactual distillation (Full FT) & Original and counterfactual data & - & - \\
KD & FDD & Feedback-driven distillation (Full FT) & Progressively add generated data & GSM8K: 49.43\% ($\uparrow$~42.53\%) & FlanT5-Large \\
KD & DLCoT & Distillation (Full FT) & High-quality data & GSM8K: 93.60\% ($\uparrow$~9.10\%) & LLaMA3.1-8B \\
KD & SKIntern & Distillation (LoRA) & Progressively simplify data & GSM8K: 33.90\% ($\uparrow$~30.80\%) & LLaMA2-7B \\
\hdashline
RL & Open-RS & GRPO (Full FT) & $\text{Blended}_1$ & AIME: 46.70\% ($\uparrow$~17.80\%) & DeepSeek-R1-Distill-Qwen-1.5B \\
RL & DeepSacleR & GRPO (Full FT) & $\text{Blended}_2$ & AIME: 43.10\% ($\uparrow$~14.20\%) & DeepSeek-R1-Distill-Qwen-1.5B \\ 
\bottomrule
\end{tabular}}
\end{table*}

\subsubsection{Distillation Transfers Reasoning Ability to Small Language Model}
\label{sec:d-transfer-reasoning-ability}

CoT-KD~\citep{magister2022teaching} first demonstrated that distillation can transfer reasoning ability from LLMs to small language models. However, due to limited capacity, small language models struggle to learn complex reasoning~\citep{li2025small}, motivating the development of more advanced strategies. Based on the optimization target, existing methods can be grouped into two directions: (1) data-focused, which improves the quality or composition of training data, and (2) model-focused, which concentrates on the distilled model itself or its generation strategy.

\paragraph{Data-focused.} MD~\citep{li2023mixed} adopts mix distillation by combining data generated with different prompting strategies (CoT and PoT) as training data, and Mix~\citep{li2025small} applies a similar strategy using a mix of long and short CoT samples. CD~\citep{feng2024teaching} enhances training diversity by mixing original data with counterfactual samples derived from it, while NAT~\citep{li2024turning} leverages negative data. DLCoT~\citep{luo2025deconstructing} improves training data quality by segmenting and simplifying long reasoning paths. SCORE~\citep{zhang2024small} enables self-correction by allowing the model to generate, identify, and refine its reasoning, using the corrected outputs for further distillation. Distill2-to-1~\citep{yu2024distilling} only retrains (input, answer) pairs as training data. The above methods rely on standard SFT, but some adopt progressive SFT. FDD~\citep{zhu2024improving} progressively adjusts data difficulty based on the small language model’s performance on LLM-generated data, while SKIntern~\citep{liao2025skintern} proposes a progressive process that removes symbolic knowledge and examples step by step, encouraging the model to internalize reasoning ability.

\paragraph{Model-focused.} PRR~\citep{zhao2024probe} distills two separate models: a probing model for retrieving relevant knowledge and a reasoning model for generating answers based on the question and retrieved content. \revise{Thinking slow, fast~\citep{paliotta2025thinking} explores distilling reasoning ability from transformer-based models into Mamba or Mamba-Transformer architectures to reduce inference cost. Similarly, M1~\citep{wang2025m1} builds on Mamba~\citep{gu2023mamba} to develop a hybrid linear RNN reasoning model that alleviates latency and memory overhead from long reasoning chains, further enhanced through RL after distillation. Additionally, works such as NSA~\citep{yuan2025native} and MoBA~\citep{lu2025moba}, which focus on lightweight architectures for general efficiency, can also be extended to improve reasoning efficiency.} Additionally, ATM~\citep{chen2024distilling} designs an adaptive mechanism that enables the student model to dynamically choose between pre-thinking (i.e., thinking before answering) and post-thinking (i.e., answering before thinking) based on question complexity.

\subsubsection{Pruning or Quantization Retain Reasoning Ability}
\label{sec:pruning-quant-retain-reasoning-ability}

Recent work~\citep{srivastava2025towards} systematically explores the impact of compression techniques like pruning and quantization on the reasoning capabilities of small language models, which shows that while quantization methods~\citep{frantar2022gptq} have minimal impact on reasoning performance, pruning approaches~\citep{li2023sparse} significantly degrade reasoning abilities. Similarly, When Reasoning Meets Compression~\citep{zhang2025reasoning} presents a comprehensive benchmark of compressed LRMs across various reasoning tasks. It also finds that quantized models retain strong reasoning performance and sometimes even surpass the original model, while aggressive pruning causes performance collapse at moderate sparsity. Furthermore, Quantization Hurts Reasoning?~\citep{liu2025quantization} systematically evaluates the impact of quantization on reasoning models. It finds that high-bit (e.g., 8-bit) quantization is nearly lossless, while low-bit settings (e.g., 4-bit) significantly degrade performance, especially on complex tasks. Interestingly, the output length of CoT reasoning remains largely unchanged, except under aggressive quantization or when using small models. \revise{Notably, the results show that on certain large models, quantization can reduce GPU memory usage by over 75\% while retaining nearly 100\% of the original performance. Meanwhile, quantized versions of large models are often more effective than standalone small models, offering advantages in both memory efficiency and performance.}

\subsubsection{Reinforcement Learning Helps Build Small Language Model}
\label{sec:build-slm-with-rl}

SLM-Foresee~\citep{srivastava2025towards} conducted a systematic study on the reasoning abilities of diverse small language models, demonstrating that small language models can exhibit strong reasoning potential. Certain models, such as the Qwen2.5 series~\citep{yang2024qwen2}, even achieve performance comparable to or surpassing some LLMs. Open-RS~\citep{dang2025reinforcement} enhanced the reasoning capability of small language models using RL with the GRPO algorithm~\citep{guo2025deepseek} and curated a high-quality mathematical reasoning dataset derived from the s1 dataset~\citep{muennighoff2025s1} and DeepScaleR dataset~\citep{luo2025deepscaler}. They further develop a cosine reward to control response length effectively. Their 1.5B model, trained on 7K samples within 24 hours on 4×A40 GPUs, achieved performance on benchmarks (e.g., AIME~24, MATH-500) that matches or surpasses models like o1-preview~\citep{o1preview2024}. SimpleRL-Zoo~\citep{zeng2025simplerl} systematically evaluated the generality of ZeroRL (i.e., an RL paradigm that enables LMs to learn long-chain reasoning with only simple rule-based rewards and no additional supervision). The study proposed several key design strategies for successful ZeroRL training: using simple correctness-based rewards, aligning data difficulty with model capacity, and employing stable RL algorithms like GRPO. Remarkably, verification behavior was observed for the first time in small language models outside the Qwen2.5 series\footnote{Most existing works focus exclusively on Qwen2.5 models, whose strong instruction following and self-reflection abilities may skew results.}, further validating the reasoning potential of small language models. Additionally, DeepScaleR\footnote{DeepScaleR is a reasoning project for small language models, code and models are available at: \url{https://github.com/agentica-project/deepscaler}}~\citep{luo2025deepscaler} leverages iterative scaling of GRPO to extend thinking length (i.e., $8\text{K} \rightarrow 16\text{K} \rightarrow 24\text{K}$), significantly improving performance on math reasoning benchmarks. The 1.5B model, DeepScaleR-1.5B-Preview, surpasses o1-Preview and achieves 43.1\% Pass@1 on AIME.

\subsection{Let Decoding More Efficient}
\label{sec:decoding-tricks}

\revise{In the previous sections, we discussed two main directions for improving reasoning efficiency.}
\revise{However, this section covers strategies to accelerate reasoning during the decoding stage.} It begins with techniques to reduce computational overhead during TTS (see Section~\ref{sec:tts-efficiency}), followed by an overview of other methods for making reasoning faster, with details provided in Section~\ref{sec:optimal-ttc}. \revise{These methods are summarized in Table~\ref{tab:faster-decoding}, showing that most methods achieve notable efficiency gains and further improve model performance without additional training.}

\begin{table*}[!t]
\centering
\caption{Overview of efficient reasoning methods in Section~\ref{sec:decoding-tricks}. The efficiency-up ratio is computed by comparing either the sampling count (S.), costs (C.), latency (L.), the correct trajectory count (T.), or FLOPs (F.). $C_1$ represents the consistency probability of the majority candidate. $C_2$ means the answer consistency within the sampling window. $C_3$ is the internal consistency via Chain-of-Embedding. $C_4$ is the probability of reaching the correct answer.}
\label{tab:faster-decoding}
\vspace{2mm}
\resizebox{0.995\linewidth}{!}{
\setlength{\tabcolsep}{0.5mm}
\begin{tabular}{llllllr} 
\toprule
\bf Type & \bf Methods & \bf Training Scheme & \bf Criteria & \bf GSM8K $\Delta$ Acc. & \bf Base Model & \bf Efficiency-up Ratio\\
\midrule
Efficient self-consistency & ASC & training-free & $C_1$ & 0.00\%  & GPT-3.5-Turbo & 1.4 - 4.3 $\times$ (S.) \\ 
Efficient self-consistency & ESC & training-free & $C_2$ & 0.00\%  & GPT-4 & 1.3 - 5.0 $\times$ (S.) \\ 
Efficient self-consistency & DSC & training-free & $C_1$ + Difficulty & $\downarrow$~0.02\%  & GPT-4 & 2.6 - 5.0 $\times$ (C.) \\ 
Efficient self-consistency & Path-Consistency & training-free & - & $\uparrow$~3.80\%  & LLaMA3-8B & 1.2 $\times$ (L.) \\ 
Efficient self-consistency & Self-Calibration & SFT (Full FT) & Confidence & $\uparrow$~2.99\% &  LLaMA3.1-8B-I  & 16.7 $\times$ (S.) \\ 
Efficient sampling & Fast Best-of-N & training-free & Reward score &  - &  -  &  
39.9 $\times$ (L.) \\ 
Efficient sampling & ST-BoN & training-free & $C_3$ & -  & - & 2.0 $\times$ (L.) \\ 
Efficient sampling & FastMCTS & training-free & $C_4$ & $\uparrow$~1.80\%  & Qwen2.5-7B & 1.1 - 3.0 $\times$ (T.) \\
Efficient sampling & Predictive-Decoding & training-free & - & $\uparrow$~0.40\% & LLaMA3-8B & - \\
Efficient sampling & $\phi$-Decoding & training-free & - & $\uparrow$~6.14\% & LLaMA3.1-8B-I & 
 2.8 $\times$ (F.) \\
Efficient sampling & Skeleton-of-Thought & training-free & - & - & - & 1.1 - 2.4 $\times$ (L.)\\
\hdashline
Other methods & AoT & training-free & - & $\uparrow$~3.00\% & GPT-4o-mini-0718 & - \\
\bottomrule
\end{tabular}}
\end{table*}

\subsubsection{Efficiency for Test-Time Scaling Strategy}
\label{sec:tts-efficiency}

While TTS strategies~\citep{snell2024scaling} have shown great promise in improving reasoning performance without modifying model weights, they often cost significant computational overhead. To make TTS more efficient, we categorize this series of works into two directions: (1) efficient sampling methods that optimize the generation process in sampling-based TTS strategies and (2) efficient self-consistency techniques that reduce the cost of consistency-based reasoning.

\paragraph{Efficient Sampling.}
\label{par:efficient-sampling}

During the sampling process, the quality of generated reasoning chains often varies, and low-quality outputs lead to substantial redundant computation. A key challenge lies in how to allocate computation more effectively. A natural solution is to terminate low-quality outputs early. Fast Best-of-N~\citep{sun2024fast} proposes speculative rejection, which halts underperforming candidates based on early-stage partial rewards. ST-BoN~\citep{wang2025sampling} adopts early consistency checks to identify and retain high-potential candidates while truncating the rest. Early path evaluation can also be applied to reasoning data synthesis. FastMCTS~\citep{li2025fastmcts} leverages MCTS to build reasoning paths while evaluating quality at each step, allowing for dynamic path adjustment. Another line of work explores predicting the future trajectory to reduce redundancy and improve overall quality. Inspired by Model Predictive Control~\citep{qin1997overview}, \citet{ma2024non} proposes Predictive-Decoding, which mitigates the myopic nature of token-level generation in CoT by simulating several future reasoning steps (i.e., foresight trajectories) to reweight the token distribution. Similarly, \citet{mendes2025language} trains a value model from the language model’s step-by-step generation dynamics to estimate the utility of intermediate reasoning states and decide whether to proceed. $\phi$-Decoding~\citep{xu2025phi} takes a step further by simulating multiple future paths at each step, clustering them to form a representative distribution and sampling the next step from this estimate.

Beyond token-level sampling, recent efforts have focused on structured sampling strategies within multi-path reasoning frameworks such as ToT and SoT. DPTS~\citep{ding2025dynamic} proposes a Dynamic Parallel Tree Search framework that parallelizes reasoning path generation and dynamically manages cache states, enabling flexible path switching without deep exploration. It also incorporates early path evaluation to prioritize promising branches. Similarly, FETCH~\citep{wang2025don} improves efficiency by merging semantically similar reasoning states to avoid redundant exploration and applying Temporal Difference (TD) learning~\citep{sutton1988learning} with $\lambda$-return to stabilize verifier scores, reducing unnecessary switching.

\paragraph{Efficient Self-Consistency.}
\label{par:efficient-self-consistency}

Self-consistency also relies on repeated sampling, which leads to substantial computational overhead. Its core challenge aligns with efficient sampling—how to allocate computation adaptively. ASC~\citep{aggarwal2023let} estimates answer confidence during sampling and stops early once sufficient confidence is observed, while ESC~\citep{li2024escape} divides the sampling process into sequential windows and stops sampling as soon as one window yields unanimous answers. DSC~\citep{wang2024make} further incorporates difficulty awareness to better adjust the sample budget per instance. RASC~\citep{wan2024reasoning} develops a similar early-stopping mechanism, terminating once sufficient high-quality samples are collected, followed by a score-weighted vote to determine the final answer. RPC~\citep{zhou2025bridging} combines self-consistency with perplexity-based estimation to accelerate convergence (i.e., the rate at which confidence estimation error for the final answer decreases with more samples). It also applies reasoning pruning to eliminate low-probability reasoning paths, reducing redundant computation. CISC~\citep{taubenfeld2025confidence} augments each sampled response with a model-predicted confidence score and performs confidence-weighted voting to improve final accuracy under the same sampling budget. Following the same idea, Self-Calibration~\citep{huang2025efficient} distills consistency signals from self-consistency into the model itself, enabling it to predict confidence scores during inference. This confidence is then used to guide early-stopping policies. Lastly, Path-Consistency~\citep{zhu2024path} extracts high-confidence reasoning prefixes from early samples and reuses them to guide future sampling, improving generation speed and answer quality.

\subsubsection{Other Methods for Making Reasoning Faster}
\label{sec:optimal-ttc}

One common approach is to decompose the original problem into sub-problems, reducing redundant token generation and skipping uninformative reasoning paths. AoT~\citep{teng2025atom} constructs a DAG to model the dependencies among initially decomposed sub-problems. It then solves the overall task by iteratively decomposing and merging sub-problems. At each step, the model only processes a simplified version of the problem, reducing unnecessary token usage, minimizing attention overhead, and avoiding memory issues caused by long contexts. DISC~\citep{light2025disc} dynamically partitions the problem into sub-steps and applies reward-based dynamic sampling and early stopping for each step to control compute costs, achieving efficient inference. AR~\citep{liu2025chaos} decomposes the reasoning process into atomic reasoning actions organized into an atomic tree and performs structured reasoning via cognitive routing (e.g., reflection, backtracking, and termination). This atomic reasoning paradigm has also proven effective in multimodal large language models (MLLMs)~\citep{xiang2025can}. \revise{SoT~\citep{ning2023skeleton} employs a two-stage decoding strategy by generating a reasoning skeleton and filling nodes in parallel. Inspired by SoT, SGD~\citep{jin2024adaptive} further builds a graph over sub-questions to capture logical dependencies and introduces difficulty-aware strategies to enable more efficient and higher-quality parallel decoding of reasoning models.}

\revise{In real-world applications, LLMs are expected to adapt their output length to input complexity, producing detailed reasoning for complex tasks and concise responses for simpler ones.} Several methods have been proposed to achieve this. TTC-Optimal Scaling~\citep{snell2024scaling} proposes a test-time compute-optimal scaling strategy that first estimates the difficulty of a prompt (i.e., either via oracle or model-predicted difficulty) and then adaptively selects different TTS strategies. For instance, on easy questions where the initial response is likely close to correct, self-verification is more efficient than multiple sampling; for complex problems, tree search with a verifier helps explore diverse reasoning paths. MRT~\citep{qu2025optimizing} further improves efficiency by introducing dense rewards based on reasoning progress (i.e., rewarding steps that increase the likelihood of reaching a correct answer) and training LLMs to progress toward solutions and avoid unnecessary computation. \revise{RSD~\citep{liao2025reward} enhances reasoning efficiency by combining a smaller draft model with a larger target model guided by a reward function.} The draft model generates candidate steps, and if the reward is high, the output is accepted; otherwise, the target model refines it. Inspired by meta-cognition~\citep{gao2024meta}, Meta-Reasoner~\citep{sui2025meta} acts as a strategic advisor to guide the reasoning process, evaluate reasoning progress, and provide high-level guidance (e.g., backtracking, restarting) based on task complexity. Additionally, SpecReason~\citep{pan2025specreason} leverages the semantic tolerance in reasoning processes by using a lightweight model to speculate intermediate steps while reserving the large model for verification and correction.

\subsection{\revise{A Supplementary: Intersections and Synergies Across Efficient Strategies.}} 
\label{sec:intersection}
\revise{Efficient reasoning strategies are not isolated, many methods combine ideas across categories to achieve better performance and flexibility. Distillation, beyond transferring reasoning capabilities, also serves as an effective means to realize latent reasoning~\citep{deng2023implicit,shen2025codi,yu2024distilling}. Its core idea further supports SFT-based methods by enabling the student model to mimic multi-step reasoning patterns~\citep{kang2024c3ot,munkhbat2025self}. Additionally, SFT and RL can be combined for adaptive reasoning. SFT is used to teach the model different answering modes, while RL helps the model learn when to switch among them based on input difficulty~\citep{fang2025thinkless,wu2025arm}.}

\section{Evaluation and Benchmark}
\label{sec:metric-benchmark}

\subsection{Metrics}
\label{sec:metrics}

Assessing reasoning efficiency requires diverse metrics reflecting computational costs and model performance (e.g., accuracy). These metrics provide insights into the trade-offs between computational efficiency and model capability, moving beyond traditional evaluation methods that solely focus on performance by incorporating additional criteria such as token count, model size, and inference latency. In the following paragraphs, we present metrics for evaluating reasoning efficiency from both general and reasoning-specific perspectives. For the general perspective, we focus on metrics related to memory, computation, and power. For the reasoning-specific perspective, we first review classic metrics used to assess reasoning capability and then discuss metrics tailored specifically for reasoning efficiency.

\subsubsection{General Perspective} 

\paragraph{Memory.} 

\begin{itemize}
    \item \textbf{Model Size} is a critical factor influencing its storage requirements and computational demands. It is commonly measured in megabytes (MB) or gigabytes (GB) and is particularly important for deployment in resource-constrained environments. Several key factors contribute to a model’s size, including parameter count, data type, and specific architectural design choices.
    \item \textbf{Memory Footprint} refers to the amount of Random Access Memory (RAM) required to run a model during training or inference. This metric is essential for understanding the model’s resource demands, particularly in environments with limited memory capacity, such as edge devices or lightweight servers. Memory is measured in units like MB or GB and is primarily determined by the model size and additional temporary data (e.g., intermediate variables).
\end{itemize}

\paragraph{Computation.} 

\begin{itemize}
    \item \textbf{Floating Point Operations (FLOPs)} measures the number of floating-point arithmetic operations a model performs during inference or training. This metric reflects a model’s computational complexity and is commonly used to assess its efficiency.
    \item \textbf{Latency (i.e., inference time)} measures the time required for an LLM to generate a response after receiving an input. This metric reflects the model’s responsiveness and is particularly important in real-world applications (e.g., chatbots) where timely outputs are essential. Latency is typically measured in seconds (s) and depends on hardware capabilities, model size, and system optimizations. Additionally, latency can be evaluated in two key ways: end-to-end latency, which measures the total time from receiving an input to producing the final output, and next-token latency, which assesses the time required to generate each token in autoregressive models. 
    \item \textbf{Throughput} measures an LLM’s efficiency by the number of tokens generated per second, typically expressed as tokens per second (TPS). It indicates overall processing capability and is crucial for batch processing or large-scale deployments. For concurrent request scenarios, throughput can be expressed as queries per second (QPS).
\end{itemize}

\paragraph{Power.}

\begin{itemize}
    \item \textbf{Power Cost} refers to the total energy consumed by an LLM throughout its lifecycle, typically measured in Watt-hours (Wh) or Joules (J). It reflects the energy usage of key hardware components such as GPUs, CPUs, and DRAM. 
    \item \textbf{Carbon Emission} measures the environmental impact of LLMs by quantifying the greenhouse gases produced during their life cycle. It is typically expressed in kilograms (kg) or tons of CO\textsubscript{2} equivalent (CO\textsubscript{2}eq) and is influenced by factors such as hardware efficiency and model runtime. Carbon emissions can be estimated as follows (see Appendix~\ref{apx:carbon} for the formula). Several tools\footnote{An online calculator: \url{https://mlco2.github.io/impact/}} are providing real-time emission tracking (e.g., CodeCarbon~\citep{schmidt2021codecarbon} and CarbonTracker~\citep{anthony2020carbontracker}) and predicting environmental costs (e.g., MLCO2 Impact~\citep{lacoste2019quantifying}).
\end{itemize}

\subsubsection{Reasoning-specific Persective}

For reasoning evaluation, several accuracy variants are used. For example, greedy accuracy measures the accuracy when decoding deterministically (i.e., selecting the most likely token at each step). Minimum-maximum spread~\citep{atil2024llm} quantifies stability by computing the accuracy gap across multiple runs. To better evaluate potential performance, the widely used Pass@k, which was initially proposed for generated code~\citep{chen2021evaluating}, has been adopted for reasoning tasks~\citep{luo2023wizardmath,yu2023metamath}. It measures the probability of obtaining at least one correct answer among $k$ independent model outputs (see Appendix~\ref{apx:pass-k} for the formula).

To capture stability, Pass$\wedge$k~\citep{yao2024tau} is proposed, which measures the probability that all $k$ generations are correct (see Appendix~\ref{apx:passk} for the formula). Pass$\wedge$k forms the basis for G-Pass@k$_\tau$~\citep{liu2024your}, which further incorporates a tolerance threshold $\tau$, requiring only a minimum proportion of correct responses among the $k$ outputs. Furthermore, to jointly assess potential and stability, mG-Pass@k$_\tau$ interpolates G-Pass@k$_\tau$ over the interval $[0.5, 1.0]$, producing a comprehensive metric (see Appendix~\ref{apx:gpass-k} for formulas).

These metrics provide a complete view of LLM reasoning performance, balancing one-shot potential with consistency across trials. Additionally, Total Agreement Rate@N (TAR@N)~\citep{atil2024llm} evaluates the consistency of a model by running it N times and measuring how often it produces identical outputs. It has two variants: TARa@N, which checks for agreement in the final answers, and TARr@N, a stricter version that requires an exact string-level match of the full outputs across runs.

To assess reasoning efficiency, token count (i.e., the number of output tokens generated by the model) is commonly used as an evaluation metric. Some studies have proposed composite metrics that integrate multiple dimensions of reasoning efficiency. CoT-Valve~\citep{ma2025cot} proposes Accuracy per Computation Unit (ACU), calculated as accuracy divided by the product of parameter count and token count, explicitly considering the trade-offs among reasoning path length, model size, and model performance. \citet{chen2024not} proposes two metrics: the outcome efficiency metric and the process efficiency metric (see Appendix~\ref{apx:outcome-process} for formulas). The outcome efficiency metric evaluates the proportion of efficient tokens (i.e., the tokens used until the first correct answer is produced) in the model-generated outputs. In contrast, the process efficiency metric assesses the diversity of reasoning paths within generated solutions.

Additionally, \citet{cuadron2025danger} introduced the overthinking score, a reliable metric explicitly designed for quantifying the degree of overthinking in LLMs. The score is obtained using an LLM-based evaluator combined with structured prompt templates. \citet{chen2024unlocking} proposed the reasoning boundary (RB) to quantify the upper limit of LLM capability in handling complex reasoning tasks (see Appendix~\ref{apx:reasoning-boundary} for the formula). \citet{wang2025thoughts} proposed the underthinking metric to evaluate whether a model prematurely abandons effective reasoning paths in incorrect responses, resulting in a large number of unproductive tokens (see Appendix~\ref{apx:underthinking-metric} for the formula).

\paragraph{Preference for Metrics: Trade-off between Performance and Efficiency.} 
In most efficient reasoning studies, performance and efficiency are typically evaluated separately—performance is measured by accuracy or Pass@k, while efficiency is assessed via token count, latency, or model size. This decoupled evaluation is simple and effective. However, some recent works have proposed unified metrics that jointly capture both aspects. For example, CoT-Valve~\citep{ma2025cot} introduces ACU, which combines parameter count, token count, and accuracy into a single metric. TALE~\citep{han2024token} proposes the optimal token budget, defined as the minimum number of tokens required to maintain correctness, and uses search algorithms to guide the model toward more efficient reasoning. Moving forward, there is a growing need for better evaluation metrics that can balance performance and efficiency more holistically and practically. O1-Pruner~\citep{luo2025o1} proposes a novel metric called the Accuracy Efficiency Score (AES), which considers both the solution length and model accuracy and penalizes accuracy degradation more than it rewards improvement (see more details in Appendix~\ref{apx:aes}).

\subsection{Datasets and Benchmarks}
\label{sec:benchmarks-datasets}

Datasets and benchmarks are crucial in evaluating language models’ reasoning capabilities and efficiency. They provide standardized protocols for assessing how well models can perform reasoning tasks under various resource constraints, such as limited computing or inference budgets. These resources cover a broad spectrum of reasoning types—including mathematical, logical, and multi-hop reasoning—enabling comprehensive evaluation across diverse domains and difficulty levels (see more details in Table~\ref{tab:datasets-benchmarks}).

\paragraph{Datasets.} To evaluate LLM reasoning ability, researchers commonly utilize developing reasoning benchmarks and datasets. Datasets are commonly categorized based on underlying reasoning types~\citep{parashar2025inference}, such as math reasoning (e.g., GSM8K~\citep{cobbe2021training}, PRM800K~\citep{lightman2023let}, MATH \& MATH-500~\citep{hendrycks2021measuring}, AIME, and AQuA~\citep{ling2017program}), logical Reasoning (e.g., ProntoQA~\citep{saparov2022language}), common sense reasoning (e.g., StrategyQA~\citep{geva2021did}, HotPotQA~\citep{yang2018hotpotqa}), algorithmic reasoning (e.g., Game of 24~\citep{yao2023tree}, Bin Packing~\citep{parashar2025inference}), and planning (e.g., BlocksWorld~\citep{valmeekam2023planning}, Rubik’s Cube~\citep{ding2023everything}, Trip Plan, and Calendar Plan~\citep{zheng2024natural}).

\paragraph{Benchmarks.} Sys2Bench~\citep{parashar2025inference} is a benchmark suite designed for evaluating LLMs, comprising 11 datasets that cover five categories of reasoning abilities (arithmetic, logical, commonsense, algorithmic, and planning). In addition to general reasoning benchmarks, several specialized benchmarks have emerged to evaluate some special situations. Overthinking Bench~\citep{cuadron2025danger} proposed a framework to assess the extent of overthinking in LLMs. Analyzing 4,018 trajectories revealed that LLMs prefer extended internal reasoning rather than environmental interactions, and it identified several undesirable behavioral patterns, such as Analysis Paralysis, Rogue Actions, and Premature Disengagement. Bag of Tricks~\citep{liu2025bag} evaluates explicitly the impact of TTC techniques on the reasoning abilities of LLMs and presents a benchmark covering six test-time optimization strategies evaluated on eight reasoning tasks. DNA Bench~\citep{hashemi2025dna} is a benchmark to assess the over-reasoning problem prevalent in current reasoning models. It comprises 150 adversarial prompts covering four key challenges (e.g., instruction adherence, hallucination avoidance, redundancy filtering, and unanswerable question recognition). DNA Bench highlights that reasoning models often produce redundant or invalid responses to simple yet misleading tasks, causing unnecessary computation and reduced accuracy.

\section{Discussions and Future Directions}
\label{sec:challenges-future-direction}

\paragraph{Efficiency Up Brings Safety Down?}
While long CoT has been shown to enhance reasoning capabilities, H-CoT~\citep{kuo2025h} reveals that LRMs can be exploited via extended CoT paths to bypass safety guardrails~\citep{feng2024unveiling}, leading to harmful outputs~\citep{li2025output}. This suggests a tension between safety and efficiency: enhancing safety requires longer, more deliberate reasoning for self-correction, which undermines efficiency, while shorter, efficient reasoning paths may skip critical safety checks. \revise{Balancing safety and efficiency remains a crucial challenge for future research in LLM reasoning. Latent reasoning offers a more structured, compact, and controllable process, making it a promising direction for reducing safety risks. Additionally, representation alignment, which constrains internal representations, may serve as a lightweight yet effective strategy for enhancing model safety.}

\paragraph{Efficient Reasoning for Multimodal Large Language Model.}
Some efficient reasoning methods can be naturally extended to the multimodal large language model (MLLM) setting. \revise{The decomposition strategy discussed in Section~\ref{sec:optimal-ttc}, which breaks complex tasks into atomic reasoning units, can also benefit multimodal reasoning~\citep{xiang2025atomicstepdecompositionenhance,hu2025sf2t}.} Similarly, latent reasoning has shown promise in MLLMs (see Heima in Section~\ref{sec:latent-reasoning}). LatentLM~\citep{sun2024multimodal} further explores this direction by unifying discrete and continuous modalities through latent language modeling. It uses a variational autoencoder (VAE) to encode continuous data into latent vectors and then applies next-token diffusion for autoregressive generation, enabling scalable and efficient multimodal generation. \revise{Additionally, efficient reasoning has been extended to typical vision tasks~\citep{wang2025pixelthink,koksal2024milchat,feng2025can,li2025tokenpacker,ouyang2023slvit,shao2025tokens}, offering valuable insights for future research on integrating structured reasoning into vision-centric multimodal applications.}

\paragraph{Break Memory Limitation.}
While long reasoning paths bring remarkable performance, they also cause severe memory issues due to long context. PENCIL~\citep{yang2025pencil} addresses this by progressively erasing outdated and unimportant reasoning steps during generation. INFTYTHINK~\citep{yan2025inftythink} adopts a segmentation strategy, breaking the reasoning path into shorter fragments and inserting concise intermediate summaries, enabling chunk-wise thinking. OMNIKV~\citep{hao2025omnikv} observes that adjacent layers share highly similar token importance distributions and thus dynamically select key tokens and reuse them across subsequent layers. MCoT~\citep{yang2024markov} models multi-step reasoning as a Markov chain, where each step depends only on the previous one, avoiding the accumulation of long historical states in the KV cache. \revise{These methods show the value of memory-efficient designs; future work should pursue lighter architectures~\citep{gu2023mamba,yuan2025native} and adaptive context management for scalable long-range reasoning.}

\paragraph{Training Efficiency.} 
Training long reasoning models remains a computationally intensive task. Recent work has aimed to improve training efficiency through both curriculum learning and RL optimization. Curriculum-based approaches, such as Light-R1~\citep{wen2025light} and FASTCURL~\citep{song2025fastcurl}, progressively increase task complexity to facilitate stable learning. Light-R1 employs curriculum SFT and multi-stage post-training, achieving strong performance with public datasets. FASTCURL extends this idea by combining curriculum RL with progressive context window extension, enabling efficient training of R1-like models even on limited hardware. On the RL front, DAPO~\citep{yu2025dapo} proposes a scalable and open-source RL system, leveraging decoupled clipping and dynamic sampling for improved training stability. AGPO~\citep{li2025adaptive} addresses critical instability in the popular GRPO~\citep{guo2025deepseek} by introducing a revised advantage estimation that mitigates zero-variance issues. Some coreset methods focus on reducing the quantity of training data. LIMO~\citep{ye2025limo} argues that complex reasoning abilities are not learned from scratch but elicited through high-quality samples. By constructing a carefully curated dataset of only 817 reasoning samples, the model trained on this data significantly outperforms those trained on nearly 100K examples. The dataset construction involves filtering out easy problems, retaining challenging ones where advanced models struggle, and performing diversity-based sampling. Similarly, s1~\citep{muennighoff2025s1} constructs a compact dataset of 1,000 examples by jointly optimizing for difficulty, diversity, and quality. \revise{Improving training efficiency through algorithmic innovations or data-centric approaches remains a promising direction with substantial room for further exploration.}

\paragraph{Opportunities in Traditional Model Compression.}
Traditional model compression techniques offer valuable opportunities for improving reasoning efficiency. Among them, distillation has demonstrated significant potential in enhancing reasoning efficiency. Distillation effectively transfers reasoning abilities from larger models to smaller ones, enabling them to achieve strong reasoning while significantly reducing costs (see Section~\ref{sec:d-transfer-reasoning-ability}). \citet{chen2025unveiling} systematically investigates three key factors that influence the effectiveness of CoT distillation: the granularity of reasoning paths, the format in which reasoning is presented, and the choice of teacher model. These insights offer practical guidance for advancing the distillation of reasoning abilities in small language models. Furthermore, distillation can play a role in other efficient reasoning directions, such as latent reasoning, where it helps compress explicit CoTs into more compact implicit reasoning paths (see Section~\ref{sec:latent-reasoning}) and SFT with variable-length CoT data (see Section~\ref{sec:sft-based-efficiency}). Distillation is a promising strategy for efficient reasoning, though there remains room for improvement. Additionally, enhancing the efficiency of the distillation process itself is also a valuable direction for future research. Beyond distillation, other model compression techniques, such as quantization and pruning, also show potential. Although preliminary pruning experiments were not promising, successful quantization suggests that model compression can maintain reasoning performance while improving efficiency in areas like memory usage.

\paragraph{Advancing Sustainability through Efficient Reasoning.}
As discussed in this work, efficient reasoning techniques contribute to optimizing the efficiency of reasoning models, reducing computational costs, and minimizing resource usage. These approaches help reduce the carbon footprint by lowering the energy requirements and supporting more environmentally friendly practices. As the use of reasoning models grows, adopting more efficient methods can play a crucial role in mitigating the environmental impact. Additionally, these efficiency improvements do not introduce significant negative effects, ensuring the benefits are realized without unintended consequences.

\paragraph{\revise{Comparison with Related Surveys.}} 
\revise{Several recent surveys have discussed reasoning models from different angles. For example, Towards Reasoning Era~\citep{chen2025reasoning} provides a comprehensive overview of long CoT reasoning, focusing primarily on reasoning performance and structure, but does not emphasize efficiency as a central concern. Some surveys~\citep{qu2025survey,sui2025stopoverthinkingsurveyefficient} center on reasoning efficiency. The former~\citep{qu2025survey} organizes methods by stages in the LLM development lifecycle (e.g., pre-training, supervised fine-tuning, reinforcement learning, and inference), offering a broad perspective across the modeling pipeline. The latter~\citep{sui2025stopoverthinkingsurveyefficient} classifies approaches based on their core technical mechanisms (e.g., model-based, output-based, and prompt-based), clearly distinguishing the underlying methodological paths. In contrast, our work focuses on how efficiency is achieved during reasoning itself, offering a goal-driven taxonomy centered around making reasoning shorter, smaller, and faster. This structured perspective helps clarify the design space of efficient reasoning and provides clearer guidance for future research.}

\paragraph{\revise{Connection between Intrinsic Efficiency Metrics and Hard Performance Metrics.}} 
\revise{In practical applications, users are primarily concerned with the efficiency that reasoning methods bring to model deployment and usage, typically measured by hard performance metrics such as time and memory. However, efficient reasoning methods often report token count rather than actual runtime. In practice, token count and latency are strongly correlated. We empirically validated this on Qwen2.5-7B using the MAHT-500 dataset, where we observed a clear positive correlation between token count and latency. The Pearson correlation coefficient was 0.9998 with a near-zero p-value, indicating a statistically significant and nearly perfect linear relationship. Meanwhile, some efficient reasoning methods employ PEFT techniques, such as LoRA, to reduce memory usage and calculation costs during the SFT or RL stages. However, this reduction applies only to the training stage and does not affect memory usage during inference or downstream deployment.}

\section{Conclusion}
\label{sec:conclusion}

In conclusion, this survey provides a comprehensive overview of efficient reasoning techniques. We categorize current efforts into three main directions—shorter, smaller, and faster—each addressing reasoning efficiency from a unique perspective: compressing reasoning chains, building small language models with strong reasoning abilities, and accelerating the decoding stage. As reasoning efficiency continues to gain traction, we believe it holds significant promise for enabling scalable and practical deployment of reasoning models across diverse applications, from real-time systems to resource-constrained environments. We hope this survey serves as a valuable foundation for future research and development in this critical and rapidly evolving field.

\section*{Acknowledgments} 

This project is supported by the Ministry of Education, Singapore, under its Academic Research Fund Tier 2 (Award Number: MOE-T2EP20122-0006).

\clearpage


\begin{thebibliography}{199}
\providecommand{\natexlab}[1]{#1}
\providecommand{\url}[1]{\texttt{#1}}
\expandafter\ifx\csname urlstyle\endcsname\relax
  \providecommand{\doi}[1]{doi: #1}\else
  \providecommand{\doi}{doi: \begingroup \urlstyle{rm}\Url}\fi

\bibitem[Aggarwal \& Welleck(2025)Aggarwal and Welleck]{aggarwal2025l1}
Pranjal Aggarwal and Sean Welleck.
\newblock L1: Controlling how long a reasoning model thinks with reinforcement learning.
\newblock \emph{arXiv preprint arXiv:2503.04697}, 2025.

\bibitem[Aggarwal et~al.(2023)Aggarwal, Madaan, Yang, et~al.]{aggarwal2023let}
Pranjal Aggarwal, Aman Madaan, Yiming Yang, et~al.
\newblock Let's sample step by step: Adaptive-consistency for efficient reasoning and coding with llms.
\newblock \emph{arXiv preprint arXiv:2305.11860}, 2023.

\bibitem[AI.(2024)]{o1preview2024}
Open AI.
\newblock Introducing openai o1-preview.
\newblock 2024.

\bibitem[Anthony et~al.(2020)Anthony, Kanding, and Selvan]{anthony2020carbontracker}
Lasse F~Wolff Anthony, Benjamin Kanding, and Raghavendra Selvan.
\newblock Carbontracker: Tracking and predicting the carbon footprint of training deep learning models.
\newblock \emph{arXiv preprint arXiv:2007.03051}, 2020.

\bibitem[Anthropic.(2025)]{claude2025}
Anthropic.
\newblock Claude 3.7 sonnet.
\newblock 2025.

\bibitem[Arora \& Zanette(2025)Arora and Zanette]{arora2025training}
Daman Arora and Andrea Zanette.
\newblock Training language models to reason efficiently.
\newblock \emph{arXiv preprint arXiv:2502.04463}, 2025.

\bibitem[Atil et~al.(2024)Atil, Chittams, Fu, Ture, Xu, and Baldwin]{atil2024llm}
Berk Atil, Alexa Chittams, Liseng Fu, Ferhan Ture, Lixinyu Xu, and Breck Baldwin.
\newblock Llm stability: A detailed analysis with some surprises.
\newblock \emph{arXiv preprint arXiv:2408.04667}, 2024.

\bibitem[Aytes et~al.(2025)Aytes, Baek, and Hwang]{aytes2025sketch}
Simon~A Aytes, Jinheon Baek, and Sung~Ju Hwang.
\newblock Sketch-of-thought: Efficient llm reasoning with adaptive cognitive-inspired sketching.
\newblock \emph{arXiv preprint arXiv:2503.05179}, 2025.

\bibitem[Besta et~al.(2024)Besta, Blach, Kubicek, Gerstenberger, Podstawski, Gianinazzi, Gajda, Lehmann, Niewiadomski, Nyczyk, et~al.]{besta2024graph}
Maciej Besta, Nils Blach, Ales Kubicek, Robert Gerstenberger, Michal Podstawski, Lukas Gianinazzi, Joanna Gajda, Tomasz Lehmann, Hubert Niewiadomski, Piotr Nyczyk, et~al.
\newblock Graph of thoughts: Solving elaborate problems with large language models.
\newblock In \emph{AAAI}, 2024.

\bibitem[Chen et~al.(2021)Chen, Tworek, Jun, Yuan, Pinto, Kaplan, Edwards, Burda, Joseph, Brockman, et~al.]{chen2021evaluating}
Mark Chen, Jerry Tworek, Heewoo Jun, Qiming Yuan, Henrique Ponde De~Oliveira Pinto, Jared Kaplan, Harri Edwards, Yuri Burda, Nicholas Joseph, Greg Brockman, et~al.
\newblock Evaluating large language models trained on code.
\newblock \emph{arXiv preprint arXiv:2107.03374}, 2021.

\bibitem[Chen et~al.(2024{\natexlab{a}})Chen, Qin, Wang, Zhou, and Che]{chen2024unlocking}
Qiguang Chen, Libo Qin, Jiaqi Wang, Jingxuan Zhou, and Wanxiang Che.
\newblock Unlocking the capabilities of thought: A reasoning boundary framework to quantify and optimize chain-of-thought.
\newblock In \emph{NeurIPS}, 2024{\natexlab{a}}.

\bibitem[Chen et~al.(2025{\natexlab{a}})Chen, Qin, Liu, Peng, Guan, Wang, Hu, Zhou, Gao, and Che]{chen2025reasoning}
Qiguang Chen, Libo Qin, Jinhao Liu, Dengyun Peng, Jiannan Guan, Peng Wang, Mengkang Hu, Yuhang Zhou, Te~Gao, and Wanxiang Che.
\newblock Towards reasoning era: A survey of long chain-of-thought for reasoning large language models.
\newblock \emph{arXiv preprint arXiv:2503.09567}, 2025{\natexlab{a}}.

\bibitem[Chen et~al.(2022)Chen, Ma, Wang, and Cohen]{chen2022program}
Wenhu Chen, Xueguang Ma, Xinyi Wang, and William~W Cohen.
\newblock Program of thoughts prompting: Disentangling computation from reasoning for numerical reasoning tasks.
\newblock \emph{arXiv preprint arXiv:2211.12588}, 2022.

\bibitem[Chen et~al.(2024{\natexlab{b}})Chen, Zhou, Liang, and Liu]{chen2024distilling}
Xiaoshu Chen, Sihang Zhou, Ke~Liang, and Xinwang Liu.
\newblock Distilling reasoning ability from large language models with adaptive thinking.
\newblock \emph{arXiv preprint arXiv:2404.09170}, 2024{\natexlab{b}}.

\bibitem[Chen et~al.(2025{\natexlab{b}})Chen, Sun, Guo, Zhang, Chen, Sun, Su, Pan, Klakow, Li, et~al.]{chen2025unveiling}
Xinghao Chen, Zhijing Sun, Wenjin Guo, Miaoran Zhang, Yanjun Chen, Yirong Sun, Hui Su, Yijie Pan, Dietrich Klakow, Wenjie Li, et~al.
\newblock Unveiling the key factors for distilling chain-of-thought reasoning.
\newblock \emph{arXiv preprint arXiv:2502.18001}, 2025{\natexlab{b}}.

\bibitem[Chen et~al.(2024{\natexlab{c}})Chen, Xu, Liang, He, Pang, Yu, Song, Liu, Zhou, Zhang, et~al.]{chen2024not}
Xingyu Chen, Jiahao Xu, Tian Liang, Zhiwei He, Jianhui Pang, Dian Yu, Linfeng Song, Qiuzhi Liu, Mengfei Zhou, Zhuosheng Zhang, et~al.
\newblock Do not think that much for 2+ 3=? on the overthinking of o1-like llms.
\newblock \emph{arXiv preprint arXiv:2412.21187}, 2024{\natexlab{c}}.

\bibitem[Chen et~al.(2024{\natexlab{d}})Chen, Lin, Sch{\"a}rli, and Zhou]{chen2023teaching}
Xinyun Chen, Maxwell Lin, Nathanael Sch{\"a}rli, and Denny Zhou.
\newblock Teaching large language models to self-debug.
\newblock In \emph{ICLR}, 2024{\natexlab{d}}.

\bibitem[Cheng \& Van~Durme(2024)Cheng and Van~Durme]{cheng2024compressed}
Jeffrey Cheng and Benjamin Van~Durme.
\newblock Compressed chain of thought: Efficient reasoning through dense representations.
\newblock \emph{arXiv preprint arXiv:2412.13171}, 2024.

\bibitem[Chuang et~al.(2024)Chuang, Zhou, Sarma, Gopalan, Boccio, Bolouki, and Hu]{chuanglearning}
Yu-Neng Chuang, Helen Zhou, Prathusha Sarma, Parikshit Gopalan, John Boccio, Sara Bolouki, and Xia Hu.
\newblock Learning to route llms with confidence tokens.
\newblock \emph{arXiv preprint arXiv:2410.13284}, 2024.

\bibitem[Chuang et~al.(2025)Chuang, Yu, Wang, Zhang, Liu, Cai, Sui, Braverman, and Hu]{chuang2025confident}
Yu-Neng Chuang, Leisheng Yu, Guanchu Wang, Lizhe Zhang, Zirui Liu, Xuanting Cai, Yang Sui, Vladimir Braverman, and Xia Hu.
\newblock Confident or seek stronger: Exploring uncertainty-based on-device llm routing from benchmarking to generalization.
\newblock \emph{arXiv preprint arXiv:2502.04428}, 2025.

\bibitem[Cobbe et~al.(2021)Cobbe, Kosaraju, Bavarian, Chen, Jun, Kaiser, Plappert, Tworek, Hilton, Nakano, et~al.]{cobbe2021training}
Karl Cobbe, Vineet Kosaraju, Mohammad Bavarian, Mark Chen, Heewoo Jun, Lukasz Kaiser, Matthias Plappert, Jerry Tworek, Jacob Hilton, Reiichiro Nakano, et~al.
\newblock Training verifiers to solve math word problems.
\newblock \emph{arXiv preprint arXiv:2110.14168}, 2021.

\bibitem[Coulom(2006)]{coulom2006efficient}
R{\'e}mi Coulom.
\newblock Efficient selectivity and backup operators in monte-carlo tree search.
\newblock In \emph{International conference on computers and games}, 2006.

\bibitem[Cuadron et~al.(2025)Cuadron, Li, Ma, Wang, Wang, Zhuang, Liu, Schroeder, Xia, Mao, et~al.]{cuadron2025danger}
Alejandro Cuadron, Dacheng Li, Wenjie Ma, Xingyao Wang, Yichuan Wang, Siyuan Zhuang, Shu Liu, Luis~Gaspar Schroeder, Tian Xia, Huanzhi Mao, et~al.
\newblock The danger of overthinking: Examining the reasoning-action dilemma in agentic tasks.
\newblock \emph{arXiv preprint arXiv:2502.08235}, 2025.

\bibitem[Cui et~al.(2024)Cui, Ma, Cao, Ye, Zhou, Liang, Chen, Lu, Yang, Liao, et~al.]{cui2024survey}
Can Cui, Yunsheng Ma, Xu~Cao, Wenqian Ye, Yang Zhou, Kaizhao Liang, Jintai Chen, Juanwu Lu, Zichong Yang, Kuei-Da Liao, et~al.
\newblock A survey on multimodal large language models for autonomous driving.
\newblock In \emph{Proceedings of the IEEE/CVF Winter Conference on Applications of Computer Vision}, 2024.

\bibitem[Cui et~al.(2025)Cui, He, Zeng, Liu, Tang, Dai, Han, Luo, Huang, Li, et~al.]{cui2025stepwise}
Yingqian Cui, Pengfei He, Jingying Zeng, Hui Liu, Xianfeng Tang, Zhenwei Dai, Yan Han, Chen Luo, Jing Huang, Zhen Li, et~al.
\newblock Stepwise perplexity-guided refinement for efficient chain-of-thought reasoning in large language models.
\newblock \emph{arXiv preprint arXiv:2502.13260}, 2025.

\bibitem[Dang \& Ngo(2025)Dang and Ngo]{dang2025reinforcement}
Quy-Anh Dang and Chris Ngo.
\newblock Reinforcement learning for reasoning in small llms: What works and what doesn't.
\newblock \emph{arXiv preprint arXiv:2503.16219}, 2025.

\bibitem[Deng et~al.(2023)Deng, Prasad, Fernandez, Smolensky, Chaudhary, and Shieber]{deng2023implicit}
Yuntian Deng, Kiran Prasad, Roland Fernandez, Paul Smolensky, Vishrav Chaudhary, and Stuart Shieber.
\newblock Implicit chain of thought reasoning via knowledge distillation.
\newblock \emph{arXiv preprint arXiv:2311.01460}, 2023.

\bibitem[Deng et~al.(2024)Deng, Choi, and Shieber]{deng2024explicit}
Yuntian Deng, Yejin Choi, and Stuart Shieber.
\newblock From explicit cot to implicit cot: Learning to internalize cot step by step.
\newblock \emph{arXiv preprint arXiv:2405.14838}, 2024.

\bibitem[Ding et~al.(2024)Ding, Liu, Fu, Song, Xie, and Zhang]{ding2024break}
Mengru Ding, Hanmeng Liu, Zhizhang Fu, Jian Song, Wenbo Xie, and Yue Zhang.
\newblock Break the chain: Large language models can be shortcut reasoners.
\newblock \emph{arXiv preprint arXiv:2406.06580}, 2024.

\bibitem[Ding et~al.(2023)Ding, Zhang, Wang, Xu, Ma, Zhang, Qin, Rajmohan, Lin, and Zhang]{ding2023everything}
Ruomeng Ding, Chaoyun Zhang, Lu~Wang, Yong Xu, Minghua Ma, Wei Zhang, Si~Qin, Saravan Rajmohan, Qingwei Lin, and Dongmei Zhang.
\newblock Everything of thoughts: Defying the law of penrose triangle for thought generation.
\newblock \emph{arXiv preprint arXiv:2311.04254}, 2023.

\bibitem[Ding et~al.(2025)Ding, Jiang, Liu, Jing, Guo, Wang, Zhang, Wang, Liu, Du, et~al.]{ding2025dynamic}
Yifu Ding, Wentao Jiang, Shunyu Liu, Yongcheng Jing, Jinyang Guo, Yingjie Wang, Jing Zhang, Zengmao Wang, Ziwei Liu, Bo~Du, et~al.
\newblock Dynamic parallel tree search for efficient llm reasoning.
\newblock \emph{arXiv preprint arXiv:2502.16235}, 2025.

\bibitem[Duan et~al.(2022)Duan, Yu, Tan, Zhu, and Tan]{duan2022survey}
Jiafei Duan, Samson Yu, Hui~Li Tan, Hongyuan Zhu, and Cheston Tan.
\newblock A survey of embodied ai: From simulators to research tasks.
\newblock \emph{IEEE Transactions on Emerging Topics in Computational Intelligence}, 6\penalty0 (2):\penalty0 230--244, 2022.

\bibitem[Fang et~al.(2023)Fang, Ma, Song, Mi, and Wang]{fang2023depgraph}
Gongfan Fang, Xinyin Ma, Mingli Song, Michael~Bi Mi, and Xinchao Wang.
\newblock Depgraph: Towards any structural pruning.
\newblock In \emph{CVPR}, 2023.

\bibitem[Fang et~al.(2024)Fang, Ma, Mi, and Wang]{fang2024isomorphic}
Gongfan Fang, Xinyin Ma, Michael~Bi Mi, and Xinchao Wang.
\newblock Isomorphic pruning for vision models.
\newblock In \emph{ECCV}, 2024.

\bibitem[Fang et~al.(2025)Fang, Ma, and Wang]{fang2025thinkless}
Gongfan Fang, Xinyin Ma, and Xinchao Wang.
\newblock Thinkless: Llm learns when to think.
\newblock \emph{arXiv preprint arXiv:2505.13379}, 2025.

\bibitem[Feng et~al.(2024{\natexlab{a}})Feng, Li, Chen, and Chen]{feng2024unveiling}
Sicheng Feng, Siyu Li, Luonan Chen, and Shengquan Chen.
\newblock Unveiling potential threats: backdoor attacks in single-cell pre-trained models.
\newblock \emph{Cell Discovery}, 10\penalty0 (1):\penalty0 122, 2024{\natexlab{a}}.

\bibitem[Feng et~al.(2024{\natexlab{b}})Feng, Tao, and Wang]{feng2024oracle}
Sicheng Feng, Keda Tao, and Huan Wang.
\newblock Is oracle pruning the true oracle?
\newblock \emph{arXiv preprint arXiv:2412.00143}, 2024{\natexlab{b}}.

\bibitem[Feng et~al.(2025)Feng, Wang, Ouyang, Kong, Song, Zhu, Wang, and Wang]{feng2025can}
Sicheng Feng, Song Wang, Shuyi Ouyang, Lingdong Kong, Zikai Song, Jianke Zhu, Huan Wang, and Xinchao Wang.
\newblock Can mllms guide me home? a benchmark study on fine-grained visual reasoning from transit maps.
\newblock \emph{arXiv preprint arXiv:2505.18675}, 2025.

\bibitem[Feng et~al.(2024{\natexlab{c}})Feng, Li, Chenglin, Chen, Yu, and Zhang]{feng2024teaching}
Tao Feng, Yicheng Li, Li~Chenglin, Hao Chen, Fei Yu, and Yin Zhang.
\newblock Teaching small language models reasoning through counterfactual distillation.
\newblock In \emph{EMNLP}, 2024{\natexlab{c}}.

\bibitem[Frantar et~al.(2023{\natexlab{a}})Frantar, Ashkboos, Hoefler, and Alistarh]{frantar-gptq}
Elias Frantar, Saleh Ashkboos, Torsten Hoefler, and Dan Alistarh.
\newblock {GPTQ}: Accurate post-training compression for generative pretrained transformers.
\newblock In \emph{ICLR}, 2023{\natexlab{a}}.

\bibitem[Frantar et~al.(2023{\natexlab{b}})Frantar, Ashkboos, Hoefler, and Alistarh]{frantar2022gptq}
Elias Frantar, Saleh Ashkboos, Torsten Hoefler, and Dan Alistarh.
\newblock Gptq: Accurate post-training quantization for generative pre-trained transformers.
\newblock In \emph{ICLR}, 2023{\natexlab{b}}.

\bibitem[Gao et~al.(2024)Gao, Xie, Mao, Wu, Xia, Mi, and Wei]{gao2024meta}
Peizhong Gao, Ao~Xie, Shaoguang Mao, Wenshan Wu, Yan Xia, Haipeng Mi, and Furu Wei.
\newblock Meta reasoning for large language models.
\newblock \emph{arXiv preprint arXiv:2406.11698}, 2024.

\bibitem[Geva et~al.(2021)Geva, Khashabi, Segal, Khot, Roth, and Berant]{geva2021did}
Mor Geva, Daniel Khashabi, Elad Segal, Tushar Khot, Dan Roth, and Jonathan Berant.
\newblock Did aristotle use a laptop? a question answering benchmark with implicit reasoning strategies.
\newblock \emph{Transactions of the Association for Computational Linguistics}, 2021.

\bibitem[Giannou et~al.(2023)Giannou, Rajput, Sohn, Lee, Lee, and Papailiopoulos]{giannou2023looped}
Angeliki Giannou, Shashank Rajput, Jy-yong Sohn, Kangwook Lee, Jason~D Lee, and Dimitris Papailiopoulos.
\newblock Looped transformers as programmable computers.
\newblock In \emph{ICML}, 2023.

\bibitem[Goel(1995)]{goel1995sketches}
Vinod Goel.
\newblock \emph{Sketches of thought}.
\newblock MIT press, 1995.

\bibitem[Gou et~al.(2024)Gou, Shao, Gong, Shen, Yang, Duan, and Chen]{gou2023critic}
Zhibin Gou, Zhihong Shao, Yeyun Gong, Yelong Shen, Yujiu Yang, Nan Duan, and Weizhu Chen.
\newblock Critic: Large language models can self-correct with tool-interactive critiquing.
\newblock In \emph{ICLR}, 2024.

\bibitem[Gray \& Neuhoff(1998)Gray and Neuhoff]{gray1998quantization}
Robert~M. Gray and David~L. Neuhoff.
\newblock Quantization.
\newblock \emph{IEEE transactions on information theory}, 1998.

\bibitem[Gu \& Dao(2024)Gu and Dao]{gu2023mamba}
Albert Gu and Tri Dao.
\newblock Mamba: Linear-time sequence modeling with selective state spaces.
\newblock In \emph{COLM}, 2024.

\bibitem[Guo et~al.(2025)Guo, Yang, Zhang, Song, Zhang, Xu, Zhu, Ma, Wang, Bi, et~al.]{guo2025deepseek}
Daya Guo, Dejian Yang, Haowei Zhang, Junxiao Song, Ruoyu Zhang, Runxin Xu, Qihao Zhu, Shirong Ma, Peiyi Wang, Xiao Bi, et~al.
\newblock Deepseek-r1: Incentivizing reasoning capability in llms via reinforcement learning.
\newblock \emph{arXiv preprint arXiv:2501.12948}, 2025.

\bibitem[Han et~al.(2016)Han, Mao, and Dally]{han2015deep}
Song Han, Huizi Mao, and William~J Dally.
\newblock Deep compression: Compressing deep neural networks with pruning, trained quantization and huffman coding.
\newblock In \emph{ICLR}, 2016.

\bibitem[Han et~al.(2024)Han, Wang, Fang, Zhao, Ma, and Chen]{han2024token}
Tingxu Han, Zhenting Wang, Chunrong Fang, Shiyu Zhao, Shiqing Ma, and Zhenyu Chen.
\newblock Token-budget-aware llm reasoning.
\newblock \emph{arXiv preprint arXiv:2412.18547}, 2024.

\bibitem[Hao et~al.(2025)Hao, Zhu, Wang, Yu, Xin, Zheng, Ren, and Guo]{hao2025omnikv}
Jitai Hao, Yuke Zhu, Tian Wang, Jun Yu, Xin Xin, Bo~Zheng, Zhaochun Ren, and Sheng Guo.
\newblock Omnikv: Dynamic context selection for efficient long-context llms.
\newblock In \emph{ICLR}, 2025.

\bibitem[Hao et~al.(2024)Hao, Sukhbaatar, Su, Li, Hu, Weston, and Tian]{hao2024training}
Shibo Hao, Sainbayar Sukhbaatar, DiJia Su, Xian Li, Zhiting Hu, Jason Weston, and Yuandong Tian.
\newblock Training large language models to reason in a continuous latent space.
\newblock \emph{arXiv preprint arXiv:2412.06769}, 2024.

\bibitem[Hashemi et~al.(2025)Hashemi, Bamgbose, Madhusudhan, Nair, Tiwari, and Yadav]{hashemi2025dna}
Masoud Hashemi, Oluwanifemi Bamgbose, Sathwik~Tejaswi Madhusudhan, Jishnu~Sethumadhavan Nair, Aman Tiwari, and Vikas Yadav.
\newblock Dna bench: When silence is smarter--benchmarking over-reasoning in reasoning llms.
\newblock \emph{arXiv preprint arXiv:2503.15793}, 2025.

\bibitem[Hendrycks et~al.(2021)Hendrycks, Burns, Kadavath, Arora, Basart, Tang, Song, and Steinhardt]{hendrycks2021measuring}
Dan Hendrycks, Collin Burns, Saurav Kadavath, Akul Arora, Steven Basart, Eric Tang, Dawn Song, and Jacob Steinhardt.
\newblock Measuring mathematical problem solving with the math dataset.
\newblock \emph{arXiv preprint arXiv:2103.03874}, 2021.

\bibitem[Hinton et~al.(2015)Hinton, Vinyals, and Dean]{hinton2015distilling}
Geoffrey Hinton, Oriol Vinyals, and Jeff Dean.
\newblock Distilling the knowledge in a neural network.
\newblock \emph{arXiv preprint arXiv:1503.02531}, 2015.

\bibitem[Hou et~al.(2025)Hou, Zhang, Ji, Liu, Qian, Andreas, and Chang]{hou2025thinkprune}
Bairu Hou, Yang Zhang, Jiabao Ji, Yujian Liu, Kaizhi Qian, Jacob Andreas, and Shiyu Chang.
\newblock Thinkprune: Pruning long chain-of-thought of llms via reinforcement learning.
\newblock \emph{arXiv preprint arXiv:2504.01296}, 2025.

\bibitem[Hu et~al.(2022)Hu, Shen, Wallis, Allen-Zhu, Li, Wang, Wang, Chen, et~al.]{hu2022lora}
Edward~J Hu, Yelong Shen, Phillip Wallis, Zeyuan Allen-Zhu, Yuanzhi Li, Shean Wang, Lu~Wang, Weizhu Chen, et~al.
\newblock Lora: Low-rank adaptation of large language models.
\newblock In \emph{ICLR}, 2022.

\bibitem[Hu et~al.(2024)Hu, Lu, Zhang, Song, Lam, and Zhang]{hu2024chain}
Hanxu Hu, Hongyuan Lu, Huajian Zhang, Yun-Ze Song, Wai Lam, and Yue Zhang.
\newblock Chain-of-symbol prompting for spatial reasoning in large language models.
\newblock In \emph{First Conference on Language Modeling}, 2024.

\bibitem[Hu et~al.(2025)Hu, Song, Feng, Luo, Yu, Chen, and Yang]{hu2025sf2t}
Yangliu Hu, Zikai Song, Na~Feng, Yawei Luo, Junqing Yu, Yi-Ping~Phoebe Chen, and Wei Yang.
\newblock Sf2t: Self-supervised fragment finetuning of video-llms for fine-grained understanding.
\newblock \emph{arXiv preprint arXiv:2504.07745}, 2025.

\bibitem[Huang et~al.(2025)Huang, Huang, Leng, Liu, and Huang]{huang2025efficient}
Chengsong Huang, Langlin Huang, Jixuan Leng, Jiacheng Liu, and Jiaxin Huang.
\newblock Efficient test-time scaling via self-calibration.
\newblock \emph{arXiv preprint arXiv:2503.00031}, 2025.

\bibitem[Jaech et~al.(2024)Jaech, Kalai, Lerer, Richardson, El-Kishky, Low, Helyar, Madry, Beutel, Carney, et~al.]{jaech2024openai}
Aaron Jaech, Adam Kalai, Adam Lerer, Adam Richardson, Ahmed El-Kishky, Aiden Low, Alec Helyar, Aleksander Madry, Alex Beutel, Alex Carney, et~al.
\newblock Openai o1 system card.
\newblock \emph{arXiv preprint arXiv:2412.16720}, 2024.

\bibitem[Jin et~al.(2024{\natexlab{a}})Jin, Luo, Cheng, Wang, Hua, Tang, Wang, and Zhang]{jin2024disentangling}
Mingyu Jin, Weidi Luo, Sitao Cheng, Xinyi Wang, Wenyue Hua, Ruixiang Tang, William~Yang Wang, and Yongfeng Zhang.
\newblock Disentangling memory and reasoning ability in large language models.
\newblock \emph{arXiv preprint arXiv:2411.13504}, 2024{\natexlab{a}}.

\bibitem[Jin et~al.(2024{\natexlab{b}})Jin, Yu, Shu, Zhao, Hua, Meng, Zhang, and Du]{jin2024impact}
Mingyu Jin, Qinkai Yu, Dong Shu, Haiyan Zhao, Wenyue Hua, Yanda Meng, Yongfeng Zhang, and Mengnan Du.
\newblock The impact of reasoning step length on large language models.
\newblock \emph{arXiv preprint arXiv:2401.04925}, 2024{\natexlab{b}}.

\bibitem[Jin et~al.(2024{\natexlab{c}})Jin, Wu, Zheng, Zhang, Lentz, Mao, Prakash, Qian, and Zhuo]{jin2024adaptive}
Shuowei Jin, Yongji Wu, Haizhong Zheng, Qingzhao Zhang, Matthew Lentz, Z~Morley Mao, Atul Prakash, Feng Qian, and Danyang Zhuo.
\newblock Adaptive skeleton graph decoding.
\newblock \emph{arXiv preprint arXiv:2402.12280}, 2024{\natexlab{c}}.

\bibitem[Kang et~al.(2024)Kang, Sun, Chen, and Zou]{kang2024c3ot}
Yu~Kang, Xianghui Sun, Liangyu Chen, and Wei Zou.
\newblock C3ot: Generating shorter chain-of-thought without compromising effectiveness.
\newblock \emph{arXiv preprint arXiv:2412.11664}, 2024.

\bibitem[Koksal \& Alatan(2025)Koksal and Alatan]{koksal2024milchat}
Aybora Koksal and Aydin~Alatan Alatan.
\newblock Milchat: Introducing chain of thought reasoning and grpo to a multimodal small language model for remote sensing.
\newblock \emph{arXiv preprint arXiv:2505.07984}, 2025.

\bibitem[Kuo et~al.(2025)Kuo, Zhang, Ding, Wang, DiValentin, Bao, Wei, Juan, Li, and Chen]{kuo2025h}
Martin Kuo, Jianyi Zhang, Aolin Ding, Qinsi Wang, Louis DiValentin, Yujia Bao, Wei Wei, Da-Cheng Juan, Hai Li, and Yiran Chen.
\newblock H-cot: Hijacking the chain-of-thought safety reasoning mechanism to jailbreak large reasoning models, including openai o1/o3, deepseek-r1, and gemini 2.0 flash thinking.
\newblock \emph{arXiv preprint arXiv:2502.12893}, 2025.

\bibitem[Lacoste et~al.(2019)Lacoste, Luccioni, Schmidt, and Dandres]{lacoste2019quantifying}
Alexandre Lacoste, Alexandra Luccioni, Victor Schmidt, and Thomas Dandres.
\newblock Quantifying the carbon emissions of machine learning.
\newblock \emph{arXiv preprint arXiv:1910.09700}, 2019.

\bibitem[LeCun et~al.(1989)LeCun, Denker, and Solla]{lecun1989optimal}
Yann LeCun, John Denker, and Sara Solla.
\newblock Optimal brain damage.
\newblock In \emph{NeurIPS}, 1989.

\bibitem[Lee et~al.(2025)Lee, Che, and Peng]{lee2025well}
Ayeong Lee, Ethan Che, and Tianyi Peng.
\newblock How well do llms compress their own chain-of-thought? a token complexity approach.
\newblock \emph{arXiv preprint arXiv:2503.01141}, 2025.

\bibitem[Li et~al.(2025{\natexlab{a}})Li, Liu, and Yang]{li2025adaptive}
Chen Li, Nazhou Liu, and Kai Yang.
\newblock Adaptive group policy optimization: Towards stable training and token-efficient reasoning.
\newblock \emph{arXiv preprint arXiv:2503.15952}, 2025{\natexlab{a}}.

\bibitem[Li et~al.(2023{\natexlab{a}})Li, Chen, Li, Wang, Li, Chen, and Zhang]{li2023mixed}
Chenglin Li, Qianglong Chen, Liangyue Li, Caiyu Wang, Yicheng Li, Zulong Chen, and Yin Zhang.
\newblock Mixed distillation helps smaller language model better reasoning.
\newblock \emph{arXiv preprint arXiv:2312.10730}, 2023{\natexlab{a}}.

\bibitem[Li et~al.(2025{\natexlab{b}})Li, Lv, Shao, Ma, Li, Zheng, Qiu, and Guo]{li2025fastmcts}
Peiji Li, Kai Lv, Yunfan Shao, Yichuan Ma, Linyang Li, Xiaoqing Zheng, Xipeng Qiu, and Qipeng Guo.
\newblock Fastmcts: A simple sampling strategy for data synthesis.
\newblock \emph{arXiv preprint arXiv:2502.11476}, 2025{\natexlab{b}}.

\bibitem[Li et~al.(2025{\natexlab{c}})Li, Yuan, Liu, Tang, Wang, Qin, Zhu, and Zhang]{li2025tokenpacker}
Wentong Li, Yuqian Yuan, Jian Liu, Dongqi Tang, Song Wang, Jie Qin, Jianke Zhu, and Lei Zhang.
\newblock Tokenpacker: Efficient visual projector for multimodal llm.
\newblock In \emph{IJCV}, 2025{\natexlab{c}}.

\bibitem[Li et~al.(2025{\natexlab{d}})Li, Li, Kosuga, and Bian]{li2025output}
Xuying Li, Zhuo Li, Yuji Kosuga, and Victor Bian.
\newblock Output length effect on deepseek-r1's safety in forced thinking.
\newblock \emph{arXiv preprint arXiv:2503.01923}, 2025{\natexlab{d}}.

\bibitem[Li et~al.(2024{\natexlab{a}})Li, Yuan, Feng, Pan, Sun, Wang, Wang, and Li]{li2024turning}
Yiwei Li, Peiwen Yuan, Shaoxiong Feng, Boyuan Pan, Bin Sun, Xinglin Wang, Heda Wang, and Kan Li.
\newblock Turning dust into gold: Distilling complex reasoning capabilities from llms by leveraging negative data.
\newblock In \emph{AAAI}, 2024{\natexlab{a}}.

\bibitem[Li et~al.(2024{\natexlab{b}})Li, Yuan, Feng, Pan, Wang, Sun, Wang, and Li]{li2024escape}
Yiwei Li, Peiwen Yuan, Shaoxiong Feng, Boyuan Pan, Xinglin Wang, Bin Sun, Heda Wang, and Kan Li.
\newblock Escape sky-high cost: Early-stopping self-consistency for multi-step reasoning.
\newblock \emph{arXiv preprint arXiv:2401.10480}, 2024{\natexlab{b}}.

\bibitem[Li et~al.(2025{\natexlab{e}})Li, Yue, Xu, Jiang, Niu, Lin, Ramasubramanian, and Poovendran]{li2025small}
Yuetai Li, Xiang Yue, Zhangchen Xu, Fengqing Jiang, Luyao Niu, Bill~Yuchen Lin, Bhaskar Ramasubramanian, and Radha Poovendran.
\newblock Small models struggle to learn from strong reasoners.
\newblock \emph{arXiv preprint arXiv:2502.12143}, 2025{\natexlab{e}}.

\bibitem[Li et~al.(2023{\natexlab{b}})Li, Niu, Zhang, Liu, Zhu, and Kang]{li2023sparse}
Yun Li, Lin Niu, Xipeng Zhang, Kai Liu, Jianchen Zhu, and Zhanhui Kang.
\newblock E-sparse: Boosting the large language model inference through entropy-based n: M sparsity.
\newblock \emph{arXiv preprint arXiv:2310.15929}, 2023{\natexlab{b}}.

\bibitem[Liao et~al.(2025{\natexlab{a}})Liao, Xu, Dong, Li, Monz, Savarese, Sahoo, and Xiong]{liao2025reward}
Baohao Liao, Yuhui Xu, Hanze Dong, Junnan Li, Christof Monz, Silvio Savarese, Doyen Sahoo, and Caiming Xiong.
\newblock Reward-guided speculative decoding for efficient llm reasoning.
\newblock \emph{arXiv preprint arXiv:2501.19324}, 2025{\natexlab{a}}.

\bibitem[Liao et~al.(2025{\natexlab{b}})Liao, He, Hao, Li, Zhang, Zhao, and Liu]{liao2025skintern}
Huanxuan Liao, Shizhu He, Yupu Hao, Xiang Li, Yuanzhe Zhang, Jun Zhao, and Kang Liu.
\newblock Skintern: Internalizing symbolic knowledge for distilling better cot capabilities into small language models.
\newblock In \emph{COLING}, 2025{\natexlab{b}}.

\bibitem[Light et~al.(2025)Light, Cheng, Yue, Oyamada, Wang, Paternain, and Chen]{light2025disc}
Jonathan Light, Wei Cheng, Wu~Yue, Masafumi Oyamada, Mengdi Wang, Santiago Paternain, and Haifeng Chen.
\newblock Disc: Dynamic decomposition improves llm inference scaling.
\newblock \emph{arXiv preprint arXiv:2502.16706}, 2025.

\bibitem[Lightman et~al.(2023)Lightman, Kosaraju, Burda, Edwards, Baker, Lee, Leike, Schulman, Sutskever, and Cobbe]{lightman2023let}
Hunter Lightman, Vineet Kosaraju, Yuri Burda, Harrison Edwards, Bowen Baker, Teddy Lee, Jan Leike, John Schulman, Ilya Sutskever, and Karl Cobbe.
\newblock Let's verify step by step.
\newblock In \emph{ICLR}, 2023.

\bibitem[Lin et~al.(2024)Lin, Tang, Tang, Yang, Chen, Wang, Xiao, Dang, Gan, and Han]{lin2023awq}
Ji~Lin, Jiaming Tang, Haotian Tang, Shang Yang, Wei-Ming Chen, Wei-Chen Wang, Guangxuan Xiao, Xingyu Dang, Chuang Gan, and Song Han.
\newblock Awq: Activation-aware weight quantization for llm compression and acceleration.
\newblock In \emph{MLSys}, 2024.

\bibitem[Ling et~al.(2017)Ling, Yogatama, Dyer, and Blunsom]{ling2017program}
Wang Ling, Dani Yogatama, Chris Dyer, and Phil Blunsom.
\newblock Program induction by rationale generation: Learning to solve and explain algebraic word problems.
\newblock \emph{arXiv preprint arXiv:1705.04146}, 2017.

\bibitem[Liu et~al.(2025{\natexlab{a}})Liu, Chao, Tan, and Liu]{liu2025bag}
Fan Liu, Wenshuo Chao, Naiqiang Tan, and Hao Liu.
\newblock Bag of tricks for inference-time computation of llm reasoning.
\newblock \emph{arXiv preprint arXiv:2502.07191}, 2025{\natexlab{a}}.

\bibitem[Liu et~al.(2025{\natexlab{b}})Liu, Zheng, Cheng, Wu, Guo, Ni, Liang, Yuan, Mao, Zhang, et~al.]{liu2025chaos}
Jinyi Liu, Yan Zheng, Rong Cheng, Qiyu Wu, Wei Guo, Fei Ni, Hebin Liang, Yifu Yuan, Hangyu Mao, Fuzheng Zhang, et~al.
\newblock From chaos to order: The atomic reasoner framework for fine-grained reasoning in large language models.
\newblock \emph{arXiv preprint arXiv:2503.15944}, 2025{\natexlab{b}}.

\bibitem[Liu et~al.(2024{\natexlab{a}})Liu, Liu, Xiao, Wang, Liu, Gao, Zhang, Zhang, and Chen]{liu2024your}
Junnan Liu, Hongwei Liu, Linchen Xiao, Ziyi Wang, Kuikun Liu, Songyang Gao, Wenwei Zhang, Songyang Zhang, and Kai Chen.
\newblock Are your llms capable of stable reasoning?
\newblock \emph{arXiv preprint arXiv:2412.13147}, 2024{\natexlab{a}}.

\bibitem[Liu et~al.(2025{\natexlab{c}})Liu, Sun, Zhang, Bai, Yu, Yu, Yuan, and Hou]{liu2025quantization}
Ruikang Liu, Yuxuan Sun, Manyi Zhang, Haoli Bai, Xianzhi Yu, Tiezheng Yu, Chun Yuan, and Lu~Hou.
\newblock Quantization hurts reasoning? an empirical study on quantized reasoning models.
\newblock \emph{arXiv preprint arXiv:2504.04823}, 2025{\natexlab{c}}.

\bibitem[Liu et~al.(2025{\natexlab{d}})Liu, Gao, Zhao, Zhang, Li, Qi, Ouyang, and Zhou]{liu2025can}
Runze Liu, Junqi Gao, Jian Zhao, Kaiyan Zhang, Xiu Li, Biqing Qi, Wanli Ouyang, and Bowen Zhou.
\newblock Can 1b llm surpass 405b llm? rethinking compute-optimal test-time scaling.
\newblock \emph{arXiv preprint arXiv:2502.06703}, 2025{\natexlab{d}}.

\bibitem[Liu et~al.(2024{\natexlab{b}})Liu, Guo, Hu, Jiayang, Zhang, Qiu, and Zhang]{liu2024can}
Tengxiao Liu, Qipeng Guo, Xiangkun Hu, Cheng Jiayang, Yue Zhang, Xipeng Qiu, and Zheng Zhang.
\newblock Can language models learn to skip steps?
\newblock \emph{arXiv preprint arXiv:2411.01855}, 2024{\natexlab{b}}.

\bibitem[Liu et~al.(2024{\natexlab{c}})Liu, Chen, Liu, Tian, and Luo]{liu2024expediting}
Tianqiao Liu, Zui Chen, Zitao Liu, Mi~Tian, and Weiqi Luo.
\newblock Expediting and elevating large language model reasoning via hidden chain-of-thought decoding.
\newblock \emph{arXiv preprint arXiv:2409.08561}, 2024{\natexlab{c}}.

\bibitem[Liu et~al.(2019)Liu, Cao, Li, Yuan, Hu, Li, and Duan]{liu2019knowledge}
Yufan Liu, Jiajiong Cao, Bing Li, Chunfeng Yuan, Weiming Hu, Yangxi Li, and Yunqiang Duan.
\newblock Knowledge distillation via instance relationship graph.
\newblock In \emph{CVPR}, 2019.

\bibitem[Lu et~al.(2025)Lu, Jiang, Liu, Du, Jiang, Hong, Liu, He, Yuan, Wang, et~al.]{lu2025moba}
Enzhe Lu, Zhejun Jiang, Jingyuan Liu, Yulun Du, Tao Jiang, Chao Hong, Shaowei Liu, Weiran He, Enming Yuan, Yuzhi Wang, et~al.
\newblock Moba: Mixture of block attention for long-context llms.
\newblock \emph{arXiv preprint arXiv:2502.13189}, 2025.

\bibitem[Luo et~al.(2023)Luo, Sun, Xu, Zhao, Lou, Tao, Geng, Lin, Chen, and Zhang]{luo2023wizardmath}
Haipeng Luo, Qingfeng Sun, Can Xu, Pu~Zhao, Jianguang Lou, Chongyang Tao, Xiubo Geng, Qingwei Lin, Shifeng Chen, and Dongmei Zhang.
\newblock Wizardmath: Empowering mathematical reasoning for large language models via reinforced evol-instruct.
\newblock \emph{arXiv preprint arXiv:2308.09583}, 2023.

\bibitem[Luo et~al.(2025{\natexlab{a}})Luo, Shen, He, Wang, Liu, Li, Tan, Cao, and Tao]{luo2025o1}
Haotian Luo, Li~Shen, Haiying He, Yibo Wang, Shiwei Liu, Wei Li, Naiqiang Tan, Xiaochun Cao, and Dacheng Tao.
\newblock O1-pruner: Length-harmonizing fine-tuning for o1-like reasoning pruning.
\newblock \emph{arXiv preprint arXiv:2501.12570}, 2025{\natexlab{a}}.

\bibitem[Luo et~al.(2025{\natexlab{b}})Luo, Tan, Wong, Shi, Tang, Roongta, Cai, Luo, Zhang, Li, et~al.]{luo2025deepscaler}
Michael Luo, Sijun Tan, Justin Wong, Xiaoxiang Shi, William~Y Tang, Manan Roongta, Colin Cai, Jeffrey Luo, Tianjun Zhang, Li~Erran Li, et~al.
\newblock Deepscaler: Surpassing o1-preview with a 1.5 b model by scaling rl.
\newblock \emph{Notion Blog}, 2025{\natexlab{b}}.

\bibitem[Luo et~al.(2025{\natexlab{c}})Luo, Song, Zhang, Liu, Wang, Chen, Su, and Zheng]{luo2025deconstructing}
Yijia Luo, Yulin Song, Xingyao Zhang, Jiaheng Liu, Weixun Wang, GengRu Chen, Wenbo Su, and Bo~Zheng.
\newblock Deconstructing long chain-of-thought: A structured reasoning optimization framework for long cot distillation.
\newblock \emph{arXiv preprint arXiv:2503.16385}, 2025{\natexlab{c}}.

\bibitem[Ma et~al.(2024)Ma, Zhao, Zhang, He, and Kong]{ma2024non}
Chang Ma, Haiteng Zhao, Junlei Zhang, Junxian He, and Lingpeng Kong.
\newblock Non-myopic generation of language models for reasoning and planning.
\newblock \emph{arXiv preprint arXiv:2410.17195}, 2024.

\bibitem[Ma et~al.(2023)Ma, Fang, and Wang]{ma2023llmpruner}
Xinyin Ma, Gongfan Fang, and Xinchao Wang.
\newblock Llm-pruner: On the structural pruning of large language models.
\newblock In \emph{NeurIPS}, 2023.

\bibitem[Ma et~al.(2025)Ma, Wan, Yu, Fang, and Wang]{ma2025cot}
Xinyin Ma, Guangnian Wan, Runpeng Yu, Gongfan Fang, and Xinchao Wang.
\newblock Cot-valve: Length-compressible chain-of-thought tuning.
\newblock \emph{arXiv preprint arXiv:2502.09601}, 2025.

\bibitem[Madaan et~al.(2023)Madaan, Tandon, Gupta, Hallinan, Gao, Wiegreffe, Alon, Dziri, Prabhumoye, Yang, et~al.]{madaan2023self}
Aman Madaan, Niket Tandon, Prakhar Gupta, Skyler Hallinan, Luyu Gao, Sarah Wiegreffe, Uri Alon, Nouha Dziri, Shrimai Prabhumoye, Yiming Yang, et~al.
\newblock Self-refine: Iterative refinement with self-feedback.
\newblock In \emph{NeurIPS}, 2023.

\bibitem[Magister et~al.(2022)Magister, Mallinson, Adamek, Malmi, and Severyn]{magister2022teaching}
Lucie~Charlotte Magister, Jonathan Mallinson, Jakub Adamek, Eric Malmi, and Aliaksei Severyn.
\newblock Teaching small language models to reason.
\newblock \emph{arXiv preprint arXiv:2212.08410}, 2022.

\bibitem[Mendes \& Ritter(2025)Mendes and Ritter]{mendes2025language}
Ethan Mendes and Alan Ritter.
\newblock Language models can self-improve at state-value estimation for better search.
\newblock \emph{arXiv preprint arXiv:2503.02878}, 2025.

\bibitem[Muennighoff et~al.(2025)Muennighoff, Yang, Shi, Li, Fei-Fei, Hajishirzi, Zettlemoyer, Liang, Cand{\`e}s, and Hashimoto]{muennighoff2025s1}
Niklas Muennighoff, Zitong Yang, Weijia Shi, Xiang~Lisa Li, Li~Fei-Fei, Hannaneh Hajishirzi, Luke Zettlemoyer, Percy Liang, Emmanuel Cand{\`e}s, and Tatsunori Hashimoto.
\newblock s1: Simple test-time scaling.
\newblock \emph{arXiv preprint arXiv:2501.19393}, 2025.

\bibitem[Munkhbat et~al.(2025)Munkhbat, Ho, Kim, Yang, Kim, and Yun]{munkhbat2025self}
Tergel Munkhbat, Namgyu Ho, Seohyun Kim, Yongjin Yang, Yujin Kim, and Se-Young Yun.
\newblock Self-training elicits concise reasoning in large language models.
\newblock \emph{arXiv preprint arXiv:2502.20122}, 2025.

\bibitem[Ning et~al.(2023)Ning, Lin, Zhou, Wang, Yang, and Wang]{ning2023skeleton}
Xuefei Ning, Zinan Lin, Zixuan Zhou, Zifu Wang, Huazhong Yang, and Yu~Wang.
\newblock Skeleton-of-thought: Prompting llms for efficient parallel generation.
\newblock \emph{arXiv preprint arXiv:2307.15337}, 2023.

\bibitem[Ong et~al.(2024)Ong, Almahairi, Wu, Chiang, Wu, Gonzalez, Kadous, and Stoica]{ong2406routellm}
Isaac Ong, Amjad Almahairi, Vincent Wu, Wei-Lin Chiang, Tianhao Wu, Joseph~E Gonzalez, M~Waleed Kadous, and Ion Stoica.
\newblock Routellm: Learning to route llms with preference data.
\newblock \emph{arXiv preprint arXiv:2406.18665}, 2024.

\bibitem[OpenAI(2024)]{openaio1}
OpenAI.
\newblock {OpenAI o1}.
\newblock \url{https://openai.com/o1/}, 2024.

\bibitem[Ouyang et~al.(2022)Ouyang, Wu, Jiang, Almeida, Wainwright, Mishkin, Zhang, Agarwal, Slama, Ray, et~al.]{ouyang2022training}
Long Ouyang, Jeffrey Wu, Xu~Jiang, Diogo Almeida, Carroll Wainwright, Pamela Mishkin, Chong Zhang, Sandhini Agarwal, Katarina Slama, Alex Ray, et~al.
\newblock Training language models to follow instructions with human feedback.
\newblock In \emph{NeurIPS}, 2022.

\bibitem[Ouyang et~al.(2023)Ouyang, Wang, Xie, Niu, Tong, Chen, and Lin]{ouyang2023slvit}
Shuyi Ouyang, Hongyi Wang, Shiao Xie, Ziwei Niu, Ruofeng Tong, Yen-Wei Chen, and Lanfen Lin.
\newblock Slvit: Scale-wise language-guided vision transformer for referring image segmentation.
\newblock In \emph{IJCAI}, 2023.

\bibitem[Paliotta et~al.(2025)Paliotta, Wang, Pagliardini, Li, Bick, Kolter, Gu, Fleuret, and Dao]{paliotta2025thinking}
Daniele Paliotta, Junxiong Wang, Matteo Pagliardini, Kevin~Y Li, Aviv Bick, J~Zico Kolter, Albert Gu, Fran{\c{c}}ois Fleuret, and Tri Dao.
\newblock Thinking slow, fast: Scaling inference compute with distilled reasoners.
\newblock \emph{arXiv preprint arXiv:2502.20339}, 2025.

\bibitem[Pan et~al.(2025)Pan, Dai, Zhang, Oliaro, Jia, and Netravali]{pan2025specreason}
Rui Pan, Yinwei Dai, Zhihao Zhang, Gabriele Oliaro, Zhihao Jia, and Ravi Netravali.
\newblock Specreason: Fast and accurate inference-time compute via speculative reasoning.
\newblock \emph{arXiv preprint arXiv:2504.07891}, 2025.

\bibitem[Parashar et~al.(2025)Parashar, Olson, Khurana, Li, Ling, Caverlee, and Ji]{parashar2025inference}
Shubham Parashar, Blake Olson, Sambhav Khurana, Eric Li, Hongyi Ling, James Caverlee, and Shuiwang Ji.
\newblock Inference-time computations for llm reasoning and planning: A benchmark and insights.
\newblock \emph{arXiv preprint arXiv:2502.12521}, 2025.

\bibitem[Pfau et~al.(2024)Pfau, Merrill, and Bowman]{pfau2024let}
Jacob Pfau, William Merrill, and Samuel~R Bowman.
\newblock Let's think dot by dot: Hidden computation in transformer language models.
\newblock In \emph{COLM}, 2024.

\bibitem[Qin \& Badgwell(1997)Qin and Badgwell]{qin1997overview}
S~Joe Qin and Thomas~A Badgwell.
\newblock An overview of industrial model predictive control technology.
\newblock In \emph{AIche symposium series}, 1997.

\bibitem[Qu et~al.(2025{\natexlab{a}})Qu, Li, Su, Sun, Yan, Liu, Cui, Liu, Liang, He, et~al.]{qu2025survey}
Xiaoye Qu, Yafu Li, Zhaochen Su, Weigao Sun, Jianhao Yan, Dongrui Liu, Ganqu Cui, Daizong Liu, Shuxian Liang, Junxian He, et~al.
\newblock A survey of efficient reasoning for large reasoning models: Language, multimodality, and beyond.
\newblock \emph{arXiv preprint arXiv:2503.21614}, 2025{\natexlab{a}}.

\bibitem[Qu et~al.(2025{\natexlab{b}})Qu, Yang, Setlur, Tunstall, Beeching, Salakhutdinov, and Kumar]{qu2025optimizing}
Yuxiao Qu, Matthew~YR Yang, Amrith Setlur, Lewis Tunstall, Edward~Emanuel Beeching, Ruslan Salakhutdinov, and Aviral Kumar.
\newblock Optimizing test-time compute via meta reinforcement fine-tuning.
\newblock \emph{arXiv preprint arXiv:2503.07572}, 2025{\natexlab{b}}.

\bibitem[Renze \& Guven(2024)Renze and Guven]{renze2024benefits}
Matthew Renze and Erhan Guven.
\newblock The benefits of a concise chain of thought on problem-solving in large language models.
\newblock In \emph{FLLM}, 2024.

\bibitem[Saparov \& He(2023)Saparov and He]{saparov2022language}
Abulhair Saparov and He~He.
\newblock Language models are greedy reasoners: A systematic formal analysis of chain-of-thought.
\newblock In \emph{ICLR}, 2023.

\bibitem[Saunshi et~al.(2025)Saunshi, Dikkala, Li, Kumar, and Reddi]{saunshi2025reasoning}
Nikunj Saunshi, Nishanth Dikkala, Zhiyuan Li, Sanjiv Kumar, and Sashank~J Reddi.
\newblock Reasoning with latent thoughts: On the power of looped transformers.
\newblock In \emph{ICLR}, 2025.

\bibitem[Schmidt et~al.(2021)Schmidt, Goyal, Joshi, Feld, Conell, Laskaris, Blank, Wilson, Friedler, and Luccioni]{schmidt2021codecarbon}
Victor Schmidt, Kamal Goyal, Aditya Joshi, Boris Feld, Liam Conell, Nikolas Laskaris, Doug Blank, Jonathan Wilson, Sorelle Friedler, and Sasha Luccioni.
\newblock Codecarbon: estimate and track carbon emissions from machine learning computing (2021).
\newblock \emph{DOI: https://doi. org/10.5281/zenodo}, 4658424, 2021.

\bibitem[Shao et~al.(2025)Shao, Tao, Zhang, Feng, Cai, Shang, You, Qin, Sui, and Wang]{shao2025tokens}
Kele Shao, Keda Tao, Kejia Zhang, Sicheng Feng, Mu~Cai, Yuzhang Shang, Haoxuan You, Can Qin, Yang Sui, and Huan Wang.
\newblock When tokens talk too much: A survey of multimodal long-context token compression across images, videos, and audios.
\newblock \emph{arXiv preprint arXiv:2507.20198}, 2025.

\bibitem[Shen et~al.(2025{\natexlab{a}})Shen, Wang, Shi, Wang, Zhao, and Gu]{shen2025efficient}
Xuan Shen, Yizhou Wang, Xiangxi Shi, Yanzhi Wang, Pu~Zhao, and Jiuxiang Gu.
\newblock Efficient reasoning with hidden thinking.
\newblock \emph{arXiv preprint arXiv:2501.19201}, 2025{\natexlab{a}}.

\bibitem[Shen et~al.(2025{\natexlab{b}})Shen, Zhang, Huang, Shi, Zhang, Yan, Wang, Wang, and Lian]{shen2025dast}
Yi~Shen, Jian Zhang, Jieyun Huang, Shuming Shi, Wenjing Zhang, Jiangze Yan, Ning Wang, Kai Wang, and Shiguo Lian.
\newblock Dast: Difficulty-adaptive slow-thinking for large reasoning models.
\newblock \emph{arXiv preprint arXiv:2503.04472}, 2025{\natexlab{b}}.

\bibitem[Shen et~al.(2025{\natexlab{c}})Shen, Yan, Zhang, Hu, Du, and He]{shen2025codi}
Zhenyi Shen, Hanqi Yan, Linhai Zhang, Zhanghao Hu, Yali Du, and Yulan He.
\newblock Codi: Compressing chain-of-thought into continuous space via self-distillation.
\newblock \emph{arXiv preprint arXiv:2502.21074}, 2025{\natexlab{c}}.

\bibitem[Snell et~al.(2024)Snell, Lee, Xu, and Kumar]{snell2024scaling}
Charlie Snell, Jaehoon Lee, Kelvin Xu, and Aviral Kumar.
\newblock Scaling llm test-time compute optimally can be more effective than scaling model parameters.
\newblock \emph{arXiv preprint arXiv:2408.03314}, 2024.

\bibitem[Song et~al.(2025)Song, Zheng, Li, Yang, Luo, Pan, and Zhang]{song2025fastcurl}
Mingyang Song, Mao Zheng, Zheng Li, Wenjie Yang, Xuan Luo, Yue Pan, and Feng Zhang.
\newblock Fastcurl: Curriculum reinforcement learning with progressive context extension for efficient training r1-like reasoning models.
\newblock \emph{arXiv preprint arXiv:2503.17287}, 2025.

\bibitem[Sprague et~al.(2024)Sprague, Yin, Rodriguez, Jiang, Wadhwa, Singhal, Zhao, Ye, Mahowald, and Durrett]{sprague2024cot}
Zayne Sprague, Fangcong Yin, Juan~Diego Rodriguez, Dongwei Jiang, Manya Wadhwa, Prasann Singhal, Xinyu Zhao, Xi~Ye, Kyle Mahowald, and Greg Durrett.
\newblock To cot or not to cot? chain-of-thought helps mainly on math and symbolic reasoning.
\newblock \emph{arXiv preprint arXiv:2409.12183}, 2024.

\bibitem[Srivastava et~al.(2025)Srivastava, Cao, and Wang]{srivastava2025towards}
Gaurav Srivastava, Shuxiang Cao, and Xuan Wang.
\newblock Towards reasoning ability of small language models.
\newblock \emph{arXiv preprint arXiv:2502.11569}, 2025.

\bibitem[Su et~al.(2025)Su, Zhu, Xu, Jiao, Tian, and Zheng]{su2025token}
DiJia Su, Hanlin Zhu, Yingchen Xu, Jiantao Jiao, Yuandong Tian, and Qinqing Zheng.
\newblock Token assorted: Mixing latent and text tokens for improved language model reasoning.
\newblock \emph{arXiv preprint arXiv:2502.03275}, 2025.

\bibitem[Sui et~al.(2025{\natexlab{a}})Sui, Chuang, Wang, Zhang, Zhang, Yuan, Liu, Wen, Chen, Hu, et~al.]{sui2025stop}
Yang Sui, Yu-Neng Chuang, Guanchu Wang, Jiamu Zhang, Tianyi Zhang, Jiayi Yuan, Hongyi Liu, Andrew Wen, Hanjie Chen, Xia Hu, et~al.
\newblock Stop overthinking: A survey on efficient reasoning for large language models.
\newblock \emph{arXiv preprint arXiv:2503.16419}, 2025{\natexlab{a}}.

\bibitem[Sui et~al.(2025{\natexlab{b}})Sui, Chuang, Wang, Zhang, Zhang, Yuan, Liu, Wen, Zhong, Chen, and Hu]{sui2025stopoverthinkingsurveyefficient}
Yang Sui, Yu-Neng Chuang, Guanchu Wang, Jiamu Zhang, Tianyi Zhang, Jiayi Yuan, Hongyi Liu, Andrew Wen, Shaochen Zhong, Hanjie Chen, and Xia Hu.
\newblock Stop overthinking: A survey on efficient reasoning for large language models.
\newblock \emph{arXiv preprint arXiv:2503.16419}, 2025{\natexlab{b}}.

\bibitem[Sui et~al.(2025{\natexlab{c}})Sui, He, Cao, Han, and Hooi]{sui2025meta}
Yuan Sui, Yufei He, Tri Cao, Simeng Han, and Bryan Hooi.
\newblock Meta-reasoner: Dynamic guidance for optimized inference-time reasoning in large language models.
\newblock \emph{arXiv preprint arXiv:2502.19918}, 2025{\natexlab{c}}.

\bibitem[Sun et~al.(2024{\natexlab{a}})Sun, Haider, Zhang, Yang, Qiu, Yin, Wang, Bartlett, and Zanette]{sun2024fast}
Hanshi Sun, Momin Haider, Ruiqi Zhang, Huitao Yang, Jiahao Qiu, Ming Yin, Mengdi Wang, Peter Bartlett, and Andrea Zanette.
\newblock Fast best-of-n decoding via speculative rejection.
\newblock In \emph{NeurIPS}, 2024{\natexlab{a}}.

\bibitem[Sun et~al.(2024{\natexlab{b}})Sun, Bao, Wang, Peng, Dong, Huang, Wang, and Wei]{sun2024multimodal}
Yutao Sun, Hangbo Bao, Wenhui Wang, Zhiliang Peng, Li~Dong, Shaohan Huang, Jianyong Wang, and Furu Wei.
\newblock Multimodal latent language modeling with next-token diffusion.
\newblock \emph{arXiv preprint arXiv:2412.08635}, 2024{\natexlab{b}}.

\bibitem[Sutton(1988)]{sutton1988learning}
Richard~S Sutton.
\newblock Learning to predict by the methods of temporal differences.
\newblock \emph{Machine learning}, 1988.

\bibitem[Tan et~al.(2025)Tan, Li, Ju, Luo, Luan, and Song]{tan2025think}
Wenhui Tan, Jiaze Li, Jianzhong Ju, Zhenbo Luo, Jian Luan, and Ruihua Song.
\newblock Think silently, think fast: Dynamic latent compression of llm reasoning chains.
\newblock \emph{arXiv preprint arXiv:2505.16552}, 2025.

\bibitem[Taubenfeld et~al.(2025)Taubenfeld, Sheffer, Ofek, Feder, Goldstein, Gekhman, and Yona]{taubenfeld2025confidence}
Amir Taubenfeld, Tom Sheffer, Eran Ofek, Amir Feder, Ariel Goldstein, Zorik Gekhman, and Gal Yona.
\newblock Confidence improves self-consistency in llms.
\newblock \emph{arXiv preprint arXiv:2502.06233}, 2025.

\bibitem[Team et~al.(2025)Team, Du, Gao, Xing, Jiang, Chen, Li, Xiao, Du, Liao, et~al.]{team2025kimi}
Kimi Team, Angang Du, Bofei Gao, Bowei Xing, Changjiu Jiang, Cheng Chen, Cheng Li, Chenjun Xiao, Chenzhuang Du, Chonghua Liao, et~al.
\newblock Kimi k1. 5: Scaling reinforcement learning with llms.
\newblock \emph{arXiv preprint arXiv:2501.12599}, 2025.

\bibitem[Teng et~al.(2025)Teng, Yu, Shi, Zhang, Wu, and Luo]{teng2025atom}
Fengwei Teng, Zhaoyang Yu, Quan Shi, Jiayi Zhang, Chenglin Wu, and Yuyu Luo.
\newblock Atom of thoughts for markov llm test-time scaling.
\newblock \emph{arXiv preprint arXiv:2502.12018}, 2025.

\bibitem[Tuo \& Wang(2025)Tuo and Wang]{tuo2025sparsessm}
Kaiwen Tuo and Huan Wang.
\newblock Sparsessm: Efficient selective structured state space models can be pruned in one-shot.
\newblock \emph{arXiv preprint arXiv:2506.09613}, 2025.

\bibitem[Valmeekam et~al.(2023)Valmeekam, Marquez, Sreedharan, and Kambhampati]{valmeekam2023planning}
Karthik Valmeekam, Matthew Marquez, Sarath Sreedharan, and Subbarao Kambhampati.
\newblock On the planning abilities of large language models-a critical investigation.
\newblock In \emph{NeurIPS}, 2023.

\bibitem[Van Den~Oord et~al.(2017)Van Den~Oord, Vinyals, et~al.]{van2017neural}
Aaron Van Den~Oord, Oriol Vinyals, et~al.
\newblock Neural discrete representation learning.
\newblock In \emph{NeurIPS}, 2017.

\bibitem[Wan et~al.(2024)Wan, Wu, Chen, and Li]{wan2024reasoning}
Guangya Wan, Yuqi Wu, Jie Chen, and Sheng Li.
\newblock Reasoning aware self-consistency: Leveraging reasoning paths for efficient llm sampling.
\newblock \emph{arXiv preprint arXiv:2408.17017}, 2024.

\bibitem[Wang et~al.(2025{\natexlab{a}})Wang, Song, Tian, Yu, Mi, Duan, Tu, Su, and Yu]{wang2025don}
Ante Wang, Linfeng Song, Ye~Tian, Dian Yu, Haitao Mi, Xiangyu Duan, Zhaopeng Tu, Jinsong Su, and Dong Yu.
\newblock Don't get lost in the trees: Streamlining llm reasoning by overcoming tree search exploration pitfalls.
\newblock \emph{arXiv preprint arXiv:2502.11183}, 2025{\natexlab{a}}.

\bibitem[Wang et~al.(2021)Wang, Qin, Zhang, and Fu]{wang2020neural}
Huan Wang, Can Qin, Yulun Zhang, and Yun Fu.
\newblock Neural pruning via growing regularization.
\newblock In \emph{ICLR}, 2021.

\bibitem[Wang et~al.(2025{\natexlab{b}})Wang, Li, Paliotta, Ritter, Rush, and Dao]{wang2025m1}
Junxiong Wang, Wen-Ding Li, Daniele Paliotta, Daniel Ritter, Alexander~M Rush, and Tri Dao.
\newblock M1: Towards scalable test-time compute with mamba reasoning models.
\newblock \emph{arXiv preprint arXiv:2504.10449}, 2025{\natexlab{b}}.

\bibitem[Wang et~al.(2024{\natexlab{a}})Wang, Ma, Feng, Zhang, Yang, Zhang, Chen, Tang, Chen, Lin, et~al.]{wang2024survey}
Lei Wang, Chen Ma, Xueyang Feng, Zeyu Zhang, Hao Yang, Jingsen Zhang, Zhiyuan Chen, Jiakai Tang, Xu~Chen, Yankai Lin, et~al.
\newblock A survey on large language model based autonomous agents.
\newblock \emph{Frontiers of Computer Science}, 18\penalty0 (6):\penalty0 186345, 2024{\natexlab{a}}.

\bibitem[Wang et~al.(2025{\natexlab{c}})Wang, Fang, Kong, Li, Xu, Yang, Li, Zhu, and Wang]{wang2025pixelthink}
Song Wang, Gongfan Fang, Lingdong Kong, Xiangtai Li, Jianyun Xu, Sheng Yang, Qiang Li, Jianke Zhu, and Xinchao Wang.
\newblock Pixelthink: Towards efficient chain-of-pixel reasoning.
\newblock \emph{arXiv preprint arXiv:2505.23727}, 2025{\natexlab{c}}.

\bibitem[Wang et~al.(2024{\natexlab{b}})Wang, Feng, Li, Yuan, Zhang, Tan, Pan, Hu, and Li]{wang2024make}
Xinglin Wang, Shaoxiong Feng, Yiwei Li, Peiwen Yuan, Yueqi Zhang, Chuyi Tan, Boyuan Pan, Yao Hu, and Kan Li.
\newblock Make every penny count: Difficulty-adaptive self-consistency for cost-efficient reasoning.
\newblock \emph{arXiv preprint arXiv:2408.13457}, 2024{\natexlab{b}}.

\bibitem[Wang et~al.(2024{\natexlab{c}})Wang, Caccia, Ostapenko, Yuan, Wang, and Sordoni]{wang2023guiding}
Xinyi Wang, Lucas Caccia, Oleksiy Ostapenko, Xingdi Yuan, William~Yang Wang, and Alessandro Sordoni.
\newblock Guiding language model reasoning with planning tokens.
\newblock In \emph{COLM}, 2024{\natexlab{c}}.

\bibitem[Wang et~al.(2022{\natexlab{a}})Wang, Wei, Schuurmans, Le, Chi, Narang, Chowdhery, and Zhou]{wang2022self}
Xuezhi Wang, Jason Wei, Dale Schuurmans, Quoc Le, Ed~Chi, Sharan Narang, Aakanksha Chowdhery, and Denny Zhou.
\newblock Self-consistency improves chain of thought reasoning in language models.
\newblock \emph{arXiv preprint arXiv:2203.11171}, 2022{\natexlab{a}}.

\bibitem[Wang et~al.(2025{\natexlab{d}})Wang, Zhang, Huang, Yang, Zhang, Huang, and Wang]{wang2025sampling}
Yiming Wang, Pei Zhang, Siyuan Huang, Baosong Yang, Zhuosheng Zhang, Fei Huang, and Rui Wang.
\newblock Sampling-efficient test-time scaling: Self-estimating the best-of-n sampling in early decoding.
\newblock \emph{arXiv preprint arXiv:2503.01422}, 2025{\natexlab{d}}.

\bibitem[Wang et~al.(2022{\natexlab{b}})Wang, Kordi, Mishra, Liu, Smith, Khashabi, and Hajishirzi]{wang2022selfinstruct}
Yizhong Wang, Yeganeh Kordi, Swaroop Mishra, Alisa Liu, Noah~A Smith, Daniel Khashabi, and Hannaneh Hajishirzi.
\newblock Self-instruct: Aligning language models with self-generated instructions.
\newblock \emph{arXiv preprint arXiv:2212.10560}, 2022{\natexlab{b}}.

\bibitem[Wang et~al.(2025{\natexlab{e}})Wang, Liu, Xu, Liang, Chen, He, Song, Yu, Li, Zhang, et~al.]{wang2025thoughts}
Yue Wang, Qiuzhi Liu, Jiahao Xu, Tian Liang, Xingyu Chen, Zhiwei He, Linfeng Song, Dian Yu, Juntao Li, Zhuosheng Zhang, et~al.
\newblock Thoughts are all over the place: On the underthinking of o1-like llms.
\newblock \emph{arXiv preprint arXiv:2501.18585}, 2025{\natexlab{e}}.

\bibitem[Wei et~al.(2022)Wei, Wang, Schuurmans, Bosma, Xia, Chi, Le, Zhou, et~al.]{wei2022chain}
Jason Wei, Xuezhi Wang, Dale Schuurmans, Maarten Bosma, Fei Xia, Ed~Chi, Quoc~V Le, Denny Zhou, et~al.
\newblock Chain-of-thought prompting elicits reasoning in large language models.
\newblock In \emph{NeurIPS}, 2022.

\bibitem[Wen et~al.(2025)Wen, Cai, Xiao, He, An, Duan, Du, Liu, Tang, Lv, et~al.]{wen2025light}
Liang Wen, Yunke Cai, Fenrui Xiao, Xin He, Qi~An, Zhenyu Duan, Yimin Du, Junchen Liu, Lifu Tang, Xiaowei Lv, et~al.
\newblock Light-r1: Curriculum sft, dpo and rl for long cot from scratch and beyond.
\newblock \emph{arXiv preprint arXiv:2503.10460}, 2025.

\bibitem[Wu et~al.(2025{\natexlab{a}})Wu, Yao, Liu, Liu, Fu, Han, Li, Zhen, Zhong, and Yuan]{wu2025unlocking}
Han Wu, Yuxuan Yao, Shuqi Liu, Zehua Liu, Xiaojin Fu, Xiongwei Han, Xing Li, Hui-Ling Zhen, Tao Zhong, and Mingxuan Yuan.
\newblock Unlocking efficient long-to-short llm reasoning with model merging.
\newblock \emph{arXiv preprint arXiv:2503.20641}, 2025{\natexlab{a}}.

\bibitem[Wu et~al.(2025{\natexlab{b}})Wu, Xie, Zhang, Chen, Zhang, Su, and Xiao]{wu2025arm}
Siye Wu, Jian Xie, Yikai Zhang, Aili Chen, Kai Zhang, Yu~Su, and Yanghua Xiao.
\newblock Arm: Adaptive reasoning model.
\newblock \emph{arXiv preprint arXiv:2505.20258}, 2025{\natexlab{b}}.

\bibitem[Wu et~al.(2025{\natexlab{c}})Wu, Sun, Li, Welleck, and Yang]{wu2024inference}
Yangzhen Wu, Zhiqing Sun, Shanda Li, Sean Welleck, and Yiming Yang.
\newblock Inference scaling laws: An empirical analysis of compute-optimal inference for problem-solving with language models.
\newblock In \emph{ICLR}, 2025{\natexlab{c}}.

\bibitem[Wu et~al.(2025{\natexlab{d}})Wu, Wang, Du, Jegelka, and Wang]{wu2025more}
Yuyang Wu, Yifei Wang, Tianqi Du, Stefanie Jegelka, and Yisen Wang.
\newblock When more is less: Understanding chain-of-thought length in llms.
\newblock \emph{arXiv preprint arXiv:2502.07266}, 2025{\natexlab{d}}.

\bibitem[Xia et~al.(2025)Xia, Li, Leong, Wang, and Li]{xia2025tokenskip}
Heming Xia, Yongqi Li, Chak~Tou Leong, Wenjie Wang, and Wenjie Li.
\newblock Tokenskip: Controllable chain-of-thought compression in llms.
\newblock \emph{arXiv preprint arXiv:2502.12067}, 2025.

\bibitem[Xiang et~al.(2025{\natexlab{a}})Xiang, Liu, Jiang, Nie, Cai, Yin, Huang, Fan, Li, Huang, Zeng, Yuan, Han, Hong, Xu, and Liang]{xiang2025atomicstepdecompositionenhance}
Kun Xiang, Zhili Liu, Zihao Jiang, Yunshuang Nie, Kaixin Cai, Yiyang Yin, Runhui Huang, Haoxiang Fan, Hanhui Li, Weiran Huang, Yihan Zeng, Yu-Jie Yuan, Jianhua Han, Lanqing Hong, Hang Xu, and Xiaodan Liang.
\newblock Can atomic step decomposition enhance the self-structured reasoning of multimodal large models?
\newblock \emph{arXiv preprint arXiv:2503.06252}, 2025{\natexlab{a}}.

\bibitem[Xiang et~al.(2025{\natexlab{b}})Xiang, Liu, Jiang, Nie, Cai, Yin, Huang, Fan, Li, Huang, et~al.]{xiang2025can}
Kun Xiang, Zhili Liu, Zihao Jiang, Yunshuang Nie, Kaixin Cai, Yiyang Yin, Runhui Huang, Haoxiang Fan, Hanhui Li, Weiran Huang, et~al.
\newblock Can atomic step decomposition enhance the self-structured reasoning of multimodal large models?
\newblock \emph{arXiv preprint arXiv:2503.06252}, 2025{\natexlab{b}}.

\bibitem[Xiao et~al.(2023)Xiao, Lin, Seznec, Wu, Demouth, and Han]{xiao2023smoothquant}
Guangxuan Xiao, Ji~Lin, Mickael Seznec, Hao Wu, Julien Demouth, and Song Han.
\newblock {S}mooth{Q}uant: Accurate and efficient post-training quantization for large language models.
\newblock In \emph{ICML}, 2023.

\bibitem[Xu et~al.(2025{\natexlab{a}})Xu, Yan, Ma, Zhao, Liu, Lin, and Wu]{xu2025phi}
Fangzhi Xu, Hang Yan, Chang Ma, Haiteng Zhao, Jun Liu, Qika Lin, and Zhiyong Wu.
\newblock $\phi$-decoding: Adaptive foresight sampling for balanced inference-time exploration and exploitation.
\newblock \emph{arXiv preprint arXiv:2503.13288}, 2025{\natexlab{a}}.

\bibitem[Xu et~al.(2025{\natexlab{b}})Xu, Hao, Zong, Wang, Zhang, Wang, Lan, Gong, Ouyang, Meng, et~al.]{xu2025towards}
Fengli Xu, Qianyue Hao, Zefang Zong, Jingwei Wang, Yunke Zhang, Jingyi Wang, Xiaochong Lan, Jiahui Gong, Tianjian Ouyang, Fanjin Meng, et~al.
\newblock Towards large reasoning models: A survey of reinforced reasoning with large language models.
\newblock \emph{arXiv preprint arXiv:2501.09686}, 2025{\natexlab{b}}.

\bibitem[Xu et~al.(2025{\natexlab{c}})Xu, Xie, Zhao, and He]{xu2025chain}
Silei Xu, Wenhao Xie, Lingxiao Zhao, and Pengcheng He.
\newblock Chain of draft: Thinking faster by writing less.
\newblock \emph{arXiv preprint arXiv:2502.18600}, 2025{\natexlab{c}}.

\bibitem[Xu et~al.(2025{\natexlab{d}})Xu, Guo, Zeng, and Miao]{xu2025softcot}
Yige Xu, Xu~Guo, Zhiwei Zeng, and Chunyan Miao.
\newblock Softcot: Soft chain-of-thought for efficient reasoning with llms.
\newblock \emph{arXiv preprint arXiv:2502.12134}, 2025{\natexlab{d}}.

\bibitem[Yan et~al.(2025)Yan, Shen, Liu, Jiang, Zhang, Shao, and Zhuang]{yan2025inftythink}
Yuchen Yan, Yongliang Shen, Yang Liu, Jin Jiang, Mengdi Zhang, Jian Shao, and Yueting Zhuang.
\newblock Inftythink: Breaking the length limits of long-context reasoning in large language models.
\newblock \emph{arXiv preprint arXiv:2503.06692}, 2025.

\bibitem[Yang et~al.(2024{\natexlab{a}})Yang, Yang, Zhang, Hui, Zheng, Yu, Li, Liu, Huang, Wei, et~al.]{yang2024qwen2}
An~Yang, Baosong Yang, Beichen Zhang, Binyuan Hui, Bo~Zheng, Bowen Yu, Chengyuan Li, Dayiheng Liu, Fei Huang, Haoran Wei, et~al.
\newblock Qwen2. 5 technical report.
\newblock \emph{arXiv preprint arXiv:2412.15115}, 2024{\natexlab{a}}.

\bibitem[Yang et~al.(2025{\natexlab{a}})Yang, Srebro, McAllester, and Li]{yang2025pencil}
Chenxiao Yang, Nathan Srebro, David McAllester, and Zhiyuan Li.
\newblock Pencil: Long thoughts with short memory.
\newblock \emph{arXiv preprint arXiv:2503.14337}, 2025{\natexlab{a}}.

\bibitem[Yang et~al.(2024{\natexlab{b}})Yang, Shen, Guo, Wang, Cao, Zhang, and Tao]{yang2024model}
Enneng Yang, Li~Shen, Guibing Guo, Xingwei Wang, Xiaochun Cao, Jie Zhang, and Dacheng Tao.
\newblock Model merging in llms, mllms, and beyond: Methods, theories, applications and opportunities.
\newblock \emph{arXiv preprint arXiv:2408.07666}, 2024{\natexlab{b}}.

\bibitem[Yang et~al.(2025{\natexlab{b}})Yang, Lin, and Yu]{yang2025think}
Junjie Yang, Ke~Lin, and Xing Yu.
\newblock Think when you need: Self-adaptive chain-of-thought learning.
\newblock \emph{arXiv preprint arXiv:2504.03234}, 2025{\natexlab{b}}.

\bibitem[Yang et~al.(2024{\natexlab{c}})Yang, Liao, and Fan]{yang2024markov}
Wen Yang, Minpeng Liao, and Kai Fan.
\newblock Markov chain of thought for efficient mathematical reasoning.
\newblock \emph{arXiv preprint arXiv:2410.17635}, 2024{\natexlab{c}}.

\bibitem[Yang et~al.(2025{\natexlab{c}})Yang, Ma, Lin, and Wei]{yang2025towards}
Wenkai Yang, Shuming Ma, Yankai Lin, and Furu Wei.
\newblock Towards thinking-optimal scaling of test-time compute for llm reasoning.
\newblock \emph{arXiv preprint arXiv:2502.18080}, 2025{\natexlab{c}}.

\bibitem[Yang et~al.(2018)Yang, Qi, Zhang, Bengio, Cohen, Salakhutdinov, and Manning]{yang2018hotpotqa}
Zhilin Yang, Peng Qi, Saizheng Zhang, Yoshua Bengio, William~W Cohen, Ruslan Salakhutdinov, and Christopher~D Manning.
\newblock Hotpotqa: A dataset for diverse, explainable multi-hop question answering.
\newblock \emph{arXiv preprint arXiv:1809.09600}, 2018.

\bibitem[Yao et~al.(2023)Yao, Yu, Zhao, Shafran, Griffiths, Cao, and Narasimhan]{yao2023tree}
Shunyu Yao, Dian Yu, Jeffrey Zhao, Izhak Shafran, Tom Griffiths, Yuan Cao, and Karthik Narasimhan.
\newblock Tree of thoughts: Deliberate problem solving with large language models.
\newblock In \emph{NeurIPS}, 2023.

\bibitem[Yao et~al.(2024)Yao, Shinn, Razavi, and Narasimhan]{yao2024tau}
Shunyu Yao, Noah Shinn, Pedram Razavi, and Karthik Narasimhan.
\newblock $\tau$-bench: A benchmark for tool-agent-user interaction in real-world domains.
\newblock \emph{arXiv preprint arXiv:2406.12045}, 2024.

\bibitem[Ye et~al.(2025)Ye, Huang, Xiao, Chern, Xia, and Liu]{ye2025limo}
Yixin Ye, Zhen Huang, Yang Xiao, Ethan Chern, Shijie Xia, and Pengfei Liu.
\newblock Limo: Less is more for reasoning.
\newblock \emph{arXiv preprint arXiv:2502.03387}, 2025.

\bibitem[Yu et~al.(2023)Yu, Jiang, Shi, Yu, Liu, Zhang, Kwok, Li, Weller, and Liu]{yu2023metamath}
Longhui Yu, Weisen Jiang, Han Shi, Jincheng Yu, Zhengying Liu, Yu~Zhang, James~T Kwok, Zhenguo Li, Adrian Weller, and Weiyang Liu.
\newblock Metamath: Bootstrap your own mathematical questions for large language models.
\newblock \emph{arXiv preprint arXiv:2309.12284}, 2023.

\bibitem[Yu et~al.(2024)Yu, Xu, Weston, and Kulikov]{yu2024distilling}
Ping Yu, Jing Xu, Jason Weston, and Ilia Kulikov.
\newblock Distilling system 2 into system 1.
\newblock \emph{arXiv preprint arXiv:2407.06023}, 2024.

\bibitem[Yu et~al.(2025{\natexlab{a}})Yu, He, Li, Zhou, Zhang, Xu, and He]{yu2025enhancing}
Qifan Yu, Zhenyu He, Sijie Li, Xun Zhou, Jun Zhang, Jingjing Xu, and Di~He.
\newblock Enhancing auto-regressive chain-of-thought through loop-aligned reasoning.
\newblock \emph{arXiv preprint arXiv:2502.08482}, 2025{\natexlab{a}}.

\bibitem[Yu et~al.(2025{\natexlab{b}})Yu, Zhang, Zhu, Yuan, Zuo, Yue, Fan, Liu, Liu, Liu, et~al.]{yu2025dapo}
Qiying Yu, Zheng Zhang, Ruofei Zhu, Yufeng Yuan, Xiaochen Zuo, Yu~Yue, Tiantian Fan, Gaohong Liu, Lingjun Liu, Xin Liu, et~al.
\newblock Dapo: An open-source llm reinforcement learning system at scale.
\newblock \emph{arXiv preprint arXiv:2503.14476}, 2025{\natexlab{b}}.

\bibitem[Yuan et~al.(2025)Yuan, Gao, Dai, Luo, Zhao, Zhang, Xie, Wei, Wang, Xiao, et~al.]{yuan2025native}
Jingyang Yuan, Huazuo Gao, Damai Dai, Junyu Luo, Liang Zhao, Zhengyan Zhang, Zhenda Xie, YX~Wei, Lean Wang, Zhiping Xiao, et~al.
\newblock Native sparse attention: Hardware-aligned and natively trainable sparse attention.
\newblock \emph{arXiv preprint arXiv:2502.11089}, 2025.

\bibitem[Zeng et~al.(2025{\natexlab{a}})Zeng, Huang, Liu, Liu, He, Ma, and He]{zeng2025simplerl}
Weihao Zeng, Yuzhen Huang, Qian Liu, Wei Liu, Keqing He, Zejun Ma, and Junxian He.
\newblock Simplerl-zoo: Investigating and taming zero reinforcement learning for open base models in the wild.
\newblock \emph{arXiv preprint arXiv:2503.18892}, 2025{\natexlab{a}}.

\bibitem[Zeng et~al.(2025{\natexlab{b}})Zeng, Cheng, Yin, Zhou, and Qiu]{zeng2025revisiting}
Zhiyuan Zeng, Qinyuan Cheng, Zhangyue Yin, Yunhua Zhou, and Xipeng Qiu.
\newblock Revisiting the test-time scaling of o1-like models: Do they truly possess test-time scaling capabilities?
\newblock \emph{arXiv preprint arXiv:2502.12215}, 2025{\natexlab{b}}.

\bibitem[Zhang et~al.(2025{\natexlab{a}})Zhang, Zhu, Sun, Luo, Qiao, Du, Zheng, Chen, and Zhang]{zhang2025lightthinker}
Jintian Zhang, Yuqi Zhu, Mengshu Sun, Yujie Luo, Shuofei Qiao, Lun Du, Da~Zheng, Huajun Chen, and Ningyu Zhang.
\newblock Lightthinker: Thinking step-by-step compression.
\newblock \emph{arXiv preprint arXiv:2502.15589}, 2025{\natexlab{a}}.

\bibitem[Zhang et~al.(2025{\natexlab{b}})Zhang, Zhang, Mitra, and Zhang]{zhang2025reasoning}
Nan Zhang, Yusen Zhang, Prasenjit Mitra, and Rui Zhang.
\newblock When reasoning meets compression: Benchmarking compressed large reasoning models on complex reasoning tasks.
\newblock \emph{arXiv preprint arXiv:2504.02010}, 2025{\natexlab{b}}.

\bibitem[Zhang et~al.(2021)Zhang, Wang, Qin, and Fu]{zhang2021learning}
Yulun Zhang, Huan Wang, Can Qin, and Yun Fu.
\newblock Learning efficient image super-resolution networks via structure-regularized pruning.
\newblock In \emph{ICLR}, 2021.

\bibitem[Zhang et~al.(2024)Zhang, Khalifa, Logeswaran, Kim, Lee, Lee, and Wang]{zhang2024small}
Yunxiang Zhang, Muhammad Khalifa, Lajanugen Logeswaran, Jaekyeom Kim, Moontae Lee, Honglak Lee, and Lu~Wang.
\newblock Small language models need strong verifiers to self-correct reasoning.
\newblock \emph{arXiv preprint arXiv:2404.17140}, 2024.

\bibitem[Zhao et~al.(2024)Zhao, Zhou, and Zhu]{zhao2024probe}
Yichun Zhao, Shuheng Zhou, and Huijia Zhu.
\newblock Probe then retrieve and reason: Distilling probing and reasoning capabilities into smaller language models.
\newblock In \emph{LREC-COLING}, 2024.

\bibitem[Zheng et~al.(2024)Zheng, Mishra, Zhang, Chen, Chen, Nova, Hou, Cheng, Le, Chi, et~al.]{zheng2024natural}
Huaixiu~Steven Zheng, Swaroop Mishra, Hugh Zhang, Xinyun Chen, Minmin Chen, Azade Nova, Le~Hou, Heng-Tze Cheng, Quoc~V Le, Ed~H Chi, et~al.
\newblock Natural plan: Benchmarking llms on natural language planning.
\newblock \emph{arXiv preprint arXiv:2406.04520}, 2024.

\bibitem[Zhou et~al.(2023)Zhou, Sch{\"a}rli, Hou, Wei, Scales, Wang, Schuurmans, Cui, Bousquet, Le, et~al.]{zhou2022least}
Denny Zhou, Nathanael Sch{\"a}rli, Le~Hou, Jason Wei, Nathan Scales, Xuezhi Wang, Dale Schuurmans, Claire Cui, Olivier Bousquet, Quoc Le, et~al.
\newblock Least-to-most prompting enables complex reasoning in large language models.
\newblock In \emph{ICLR}, 2023.

\bibitem[Zhou et~al.(2025)Zhou, Yuhao, Li, Yao, Guo, Ma, and Li]{zhou2025bridging}
Zhi Zhou, Tan Yuhao, Zenan Li, Yuan Yao, Lan-Zhe Guo, Xiaoxing Ma, and Yu-Feng Li.
\newblock Bridging internal probability and self-consistency for effective and efficient llm reasoning.
\newblock \emph{arXiv preprint arXiv:2502.00511}, 2025.

\bibitem[Zhu et~al.(2024{\natexlab{a}})Zhu, Shen, Zhao, and Zou]{zhu2024path}
Jiace Zhu, Yingtao Shen, Jie Zhao, and An~Zou.
\newblock Path-consistency: Prefix enhancement for efficient inference in llm.
\newblock \emph{arXiv preprint arXiv:2409.01281}, 2024{\natexlab{a}}.

\bibitem[Zhu et~al.(2024{\natexlab{b}})Zhu, Li, Ma, and Wang]{zhu2024improving}
Xunyu Zhu, Jian Li, Can Ma, and Weiping Wang.
\newblock Improving mathematical reasoning capabilities of small language models via feedback-driven distillation.
\newblock \emph{arXiv preprint arXiv:2411.14698}, 2024{\natexlab{b}}.

\end{thebibliography}
\bibliographystyle{tmlr}


\appendix

\section{Appendix}

\subsection{Details for Model Compression}
\label{apx:model-compression-details}

\paragraph{Quantization.}
Quantization improves model efficiency and reduces memory usage by lowering the bit precision of parameters. It is typically categorized into post-training quantization (PTQ) and quantization-aware training (QAT), distinguished by whether retraining is involved. PTQ applies quantization directly to a pre-trained model, while QAT includes a retraining stage to mitigate quantization-induced errors. Quantization can target weights, activations, or both. Advanced methods such as GPTQ~\citep{frantar-gptq}, AWQ~\citep{lin2023awq}, and SmoothQuant~\citep{xiao2023smoothquant} further enhance quantization for large language models by reducing activation outliers and minimizing calibration errors.

\paragraph{Pruning.}
Pruning reduces model size and inference latency by eliminating redundant or less important parameters. It can be broadly categorized into unstructured pruning, structured pruning, and semi-structured pruning. Unstructured pruning removes individual weights based on certain criteria, such as magnitude. While it achieves high sparsity, it is often less hardware-friendly due to irregular sparsity patterns. Structured pruning eliminates entire units such as neurons, channels, or attention heads, leading to more regular sparsity patterns that are easier to accelerate in practice. Semi-structured pruning strikes a balance between the two, applying constraints such as N:M sparsity, where only a fixed number of weights are retained in each block. This enables efficient execution on specialized hardware.
Recent works (e.g., LLM-Pruner, DepGraph)~\citep{ma2023llmpruner,fang2024isomorphic,fang2023depgraph,feng2024oracle}, and methods based on importance scores and gradient sensitivity~\citep{wang2020neural,zhang2021learning,tuo2025sparsessm} have significantly improved the effectiveness and usability of pruning for large models.

\paragraph{Knowledge Distillation.}
Knowledge Distillation (KD) transfers the behavior of a large, well-performing teacher model to a smaller student model by aligning output distributions (e.g., logits or soft labels), intermediate representations, or attention patterns. KD approaches can be categorized as black-box or white-box, depending on whether the student has access only to the teacher’s outputs or to internal states as well. Variants like Self-Instruct KD~\citep{wang2022selfinstruct} enable the student to inherit reasoning abilities and generalization skills in more flexible settings.

\subsection{Overthinking Example}
\label{apx:overthinking-example}

We provide an example to show the overthinking problem: for the simple question ``What is the answer of 1 plus 2?'', DeepSeek R1 generates a lengthy response containing 619 tokens, whereas the ideal answer ``The answer is 3.'' requires only 17 tokens. This highlights the inefficiency brought by overthinking in current models, especially when solving simple problems.

\begin{tcolorbox}[deepseekbox]
Okay, so I need to figure out what 1 plus 2 is. Let me start by recalling basic addition. When you add two numbers together, you’re combining their values. So, if I have 1 apple and someone gives me 2 more apples, how many apples do I have in total?

Let me count them out. Starting with 1, then adding another 2. So, 1... then 2, 3. Wait, does that mean 1 plus 2 equals 3? Hmm, that seems right. Let me check another way to make sure I’m not making a mistake. Maybe using my fingers. Hold up one finger on my left hand for the number 1, then two fingers on my right hand for the number 2. Now, if I count all the fingers I’m holding up, that’s 1, 2, 3. Yep, that’s three fingers. So 1 plus 2 must be 3.

Alternatively, I can think of the number line. Starting at 1, then moving 2 steps to the right. So from 1, moving one step to 2, then another step to 3. That lands me on 3. So that also gives me 3.

Wait, is there another method to verify? Maybe using objects. If I have 1 book and add 2 more books, the total should be 3 books. Let me visualize: one book here, two there. Combine them, count them all: 1, 2, 3. Yep, three books.

I guess all these methods point to the same answer. But just to be thorough, maybe I can use the basic addition facts I learned. The sum of 1 and 2 is one of the fundamental arithmetic facts. From what I remember, 1 + 2 is indeed 3. There’s no carrying over or anything complicated here because both numbers are single-digit.

Another way to look at it is through set theory. If I have a set with 1 element and another set with 2 elements, the union of these sets will have 1 + 2 = 3 elements, provided there’s no overlap. So, assuming the sets are disjoint, the total number of elements is 3.

Also, in terms of equations, if I write it out: 1 + 2 = x. Solving for x, I can subtract 1 from both sides, but that might complicate things. Alternatively, just recognizing that addition is commutative, so 1 + 2 is the same as 2 + 1, which is 3.

I think I’ve checked it multiple ways: counting on fingers, using a number line, set theory, and basic arithmetic facts. All confirm that 1 plus 2 equals 3. I don’t see any reason to doubt this result. It’s straightforward and consistent across different methods. So the answer must be 3.
\end{tcolorbox}

\begin{tcolorbox}[expectedbox]
The answer is 3.
\end{tcolorbox}

\subsection{\revise{Latency Comparison of Efficient Reasoning Methods}}
\label{apx:latency-comparison}

\revise{Table~\ref{tab:gsm8k-all-category} summarizes representative efficient reasoning methods on GSM8K across different categories, providing a practical overview of efficient reasoning approaches for users.}

\begin{table*}[!t]
\centering
\caption{Overview of efficient reasoning methods on GSM8K. The speedup ratio is computed mainly through latency comparison, except for Self-Calibration, where sampling count (S.) is used as a proxy.}
\label{tab:gsm8k-all-category}
\vspace{2mm}
\resizebox{0.995\linewidth}{!}{
\setlength{\tabcolsep}{0.5mm}
\begin{tabular}{lllllr}
\toprule
\bf Category / Type & \bf Methods & \bf Training Scheme & \bf Accuracy & \bf Base Model & \bf Speedup \\
\midrule
Shorter / Routing & Self-REF &  SFT (LoRA) & \bf 81.60\% & LLaMA3-8B-I & 1.3 $\times$   \\
Smaller / KD & SKIntern & Distillation (LoRA) &  62.50\%  & LLaMA3-8B-I & - \\
Faster / Efficient self-consistency & Path-Consistency & Training-free & 67.80\%  & LLaMA3-8B-I & 1.2 $\times$  \\ 
\hdashline
Shorter / SFT  & CoT-Valve & Progressive SFT (LoRA) & 87.30\%  & LLaMA3.1-8B-I & 1.7 $\times$ \\ 
Shorter / SFT  & TokenSkip &  SFT (LoRA) & 78.20\%   & LLaMA3.1-8B-I  & 1.7 - 1.8 $\times$   \\ 
Shorter / SFT  & TALE-PT &  SFT (LoRA) & 78.57\%   & LLaMA3.1-8B-I  & 1.7 $\times$   \\ 
Shorter / Latent reasoning & SoftCoT &  SFT (Freeze FT) & 81.03\%    & LLaMA3.1-8B-I & 4.0 - 5.0 $\times$   \\
Shorter / Latent reasoning & LightThinker &  SFT (Full FT) & 88.25\%    & LLaMA3.1-8B-I & up to 1.4 $\times$  \\
Shorter / Latent reasoning & Token Assorted & SFT (Full FT) & 84.10\%    & LLaMA3.1-8B-I & 1.2 $\times$ \\
Smaller / KD & Mix & Mixed distillation (Full FT \& LoRA)   & 81.40\%  & LLaMA3.1-8B-I & -\\
Smaller / KD & DLCoT & Distillation (Full FT) &  \bf 93.60\% & LLaMA3.1-8B-I & - \\
Faster / Efficient sampling & $\phi$-Decoding & Training-free &  86.58\% & LLaMA3.1-8B-I & 2.8 $\times$ \\
Faster / Efficient self-consistency & Self-Calibration & SFT (Full FT) & 80.43\%  &  LLaMA3.1-8B-I  & 16.7 $\times$ (S.) \\ 
\bottomrule
\end{tabular}}
\end{table*}

\subsection{Metric Formulas}
\label{sec:metric-detail}

\subsubsection{Carbon Emission}
\label{apx:carbon}

\begin{equation}
\underset{(\text{kgCO}_2\text{eq})}{\text{Carbon Emission}} = 
\underset{(\text{kWh})}{\text{Energy Consumption}} \times
 \underset{(\text{gCO}_2\text{eq/kWh})}{\text{Carbon Intensity}}
\end{equation}

\subsubsection{Pass@k}
\label{apx:pass-k}

\begin{equation}
\text{Pass@}k = 1 - \mathbb{E}_{\text{task}}\left[ \frac{{n-c \choose k}}{{n \choose k}} \right]
\end{equation}
where $n$ is the number of sampled outputs and $c$ is the number of correct ones.

\subsubsection{Pass$\wedge$k}
\label{apx:passk}

\begin{equation}
Pass \wedge k = \mathbb{E}_{\text{task}}\left[ \frac{{c \choose k}}{{n \choose k}} \right]
\end{equation}
where $n$ is the number of sampled outputs and $c$ is the number of correct ones.

\subsubsection{G-Pass@k}
\label{apx:gpass-k}

\begin{equation}
\text{G-Pass@}k_{\tau} = \mathbb{E}_{\text{task}} \left[ \sum_{j=\lceil \tau k \rceil}^{c} \frac{{c \choose j} {n - c \choose k - j}}{{n \choose k}} \right]
\end{equation}
where $n$ is the number of sampled outputs, $c$ is the number of correct ones, and 
$\tau$ is a tolerance threshold that represents the minimum proportion of correct responses among the $k$ outputs.

\begin{equation}
\text{mG-Pass@}k_{\tau} = \frac{2}{k} \sum_{i=\lceil 0.5k \rceil + 1}^{k} \text{G-Pass@}k_{\frac{i}{k}}
\end{equation}

\subsubsection{Outcome and Process Efficiency Metric}
\label{apx:outcome-process}

\paragraph{Outcome Efficiency Metric:}
\begin{equation}
\xi_{O} = \frac{1}{N}\sum_{i=1}^{N}\sigma_{i}\frac{\hat{T}_{i}}{T_{i}}
\end{equation}
where \( N \) is the number of instances, \( T_i \) denotes the total number of tokens generated for instance \( i \), \( \hat{T}_i \) is the number of tokens until the first correct answer, and \( \sigma_i \) indicates correctness:
\[
\sigma_i = \begin{cases}
1, & \text{if at least one solution is correct} \\
0, & \text{otherwise}
\end{cases}
\]

\paragraph{Process Efficiency Metric:}
\begin{equation}
\xi_{P} = \frac{1}{N}\sum_{i=1}^{N}\frac{D_i}{T_i}
\end{equation}
where \( D_i \) represents tokens contributing to solution diversity, defined as:
\[
D_i = \sum_{m=1}^{M}\tau_i^{m}T_i^{m}
\]
where \( T_i^{m} \) is the token count of the \( m \)-th solution for instance \( i \), and \( \tau_i^{m} \) denotes whether the solution introduces a new reasoning strategy:
\[
\tau_i^{m} = \begin{cases}
1, & \text{if solution } m \text{ is distinct in reasoning} \\
0, & \text{otherwise}
\end{cases}
\]

\subsubsection{Reasoning Boundary (RB)}
\label{apx:reasoning-boundary}

\begin{equation}
B_{Acc=K_1} (t|m) = \sup_d \left\{ d \mid \text{Acc}(t|d, m) = K_1 \right\}
\end{equation}
where $t$ denotes a specific reasoning task, $m$ represents the evaluated language model, $d$ indicates the difficulty level of the task, $\text{Acc}(t|d, m)$ is the accuracy of model $m$ on task $t$ with difficulty $d$, $K_1$ is a predefined accuracy threshold, $\sup$ denotes the supremum (least upper bound) over the set of difficulty levels satisfying the accuracy condition.

\subsubsection{Underthinking Metric}
\label{apx:underthinking-metric}

\begin{equation}
\xi_{\text{UT}} = \frac{1}{N} \sum_{i=1}^N \left(1 - \frac{\hat{T}_i}{T_i} \right)
\end{equation}
where $N$ is the number of incorrect response instances in the test set, $T_i$ is the total number of tokens in the $i$-th incorrect response, $\hat{T}_i$ is the number of tokens from the beginning of the $i$-th response up to and including the first correct thought.

\subsubsection{Accuracy Efficiency Score}
\label{apx:aes}

\begin{align*}
\Delta \text{Length} &= \frac{\text{Length}_{\text{baseline}} - \text{Length}_{\text{model}}}{\text{Length}_{\text{baseline}}}, \\
\Delta \text{Acc} &= \frac{\text{Acc}_{\text{model}} - \text{Acc}_{\text{baseline}}}{\text{Acc}_{\text{baseline}}}
\end{align*}

Then, the AES is computed as:

\[
\text{AES} =
\begin{cases}
\alpha \cdot \Delta \text{Length} + \beta \cdot |\Delta \text{Acc}|, & \text{if } \Delta \text{Acc} \geq 0 \\
\alpha \cdot \Delta \text{Length} - \gamma \cdot |\Delta \text{Acc}|, & \text{if } \Delta \text{Acc} < 0
\end{cases}
\]

where $\alpha > 0$, $\beta > 0$, and $\gamma > 0$ are weighting factors. The default values $\alpha=1$, $\beta=3$, and $\gamma=5$ are used to emphasize penalizing accuracy drop more heavily than rewarding accuracy improvement.

\subsection{Complete List of Datasets and Benchmarks}
\label{apx:dataset-benchmark-table}

\begin{table*}[!t]
\centering
\caption{Full List of Datasets and Benchmarks.}
\label{tab:datasets-benchmarks}
\vspace{2mm}
\resizebox{0.995\linewidth}{!}{
\setlength{\tabcolsep}{6mm}
\begin{tabular}{llll}
\toprule
Type & Name & Task / Target & Source \\
\midrule
\emph{Datasets} & GSM8K & Math &  \href{https://huggingface.co/datasets/openai/gsm8k}{HuggingFace Dataset} \\ 
& MATH \& MATH-500 & Math & \href{https://huggingface.co/datasets/HuggingFaceH4/MATH-500}{HuggingFace Dataset}  \\ 
& AIME & Math & \href{https://huggingface.co/datasets/HuggingFaceH4/aime_2024}{HuggingFace Dataset}  \\ 
& AMC & Math & \href{https://huggingface.co/datasets/math-ai/amc23}{HuggingFace Dataset}  \\ 
& AQuA & Math & \href{https://huggingface.co/datasets/deepmind/aqua_rat}{HuggingFace Dataset}  \\ 
& ProntoQA & Logical & \href{https://github.com/asaparov/prontoqa}{GitHub}  \\ 
& StrategyQA & Common sense &  \href{https://huggingface.co/datasets/ChilleD/StrategyQA}{HuggingFace Dataset} \\ 
& HotPotQA & Common sense & \href{https://huggingface.co/datasets/hotpotqa/hotpot_qa}{HuggingFace Dataset}  \\ 
& Game of 24 & Algorithmic &  \href{https://github.com/princeton-nlp/tree-of-thought-llm}{GitHub} \\ 
& Bin Packing & Algorithmic &  \href{https://github.com/divelab/sys2bench}{GitHub} \\ 
& BlocksWorld & Planning &  \href{https://huggingface.co/datasets/chiayewken/blocksworld}{HuggingFace Dataset} \\ 
& Rubik’s Cube & Planning &  \href{https://github.com/divelab/sys2bench}{GitHub} \\ 
& Trip Plan & Planning & \href{https://github.com/divelab/sys2bench}{GitHub}  \\ 
& Calendar Plan & Planning & \href{https://github.com/divelab/sys2bench}{GitHub}  \\ 
\hdashline
\emph{Benchmarks} & Sys2Bench & General reasoning &  \href{https://github.com/divelab/sys2bench}{GitHub}  \\
& Overthinking Bench & Overthinking & \href{https://github.com/AlexCuadron/ThinkingAgent}{GitHub} \\
& Bag of Tricks & Test-time computation (TTC) & \href{https://github.com/usail-hkust/benchmark_inference_time_computation_LLM}{GitHub} \\
& DNA Bench & Over-reasoning & - \\
\bottomrule
\end{tabular}}
\end{table*}

A complete list of the datasets and benchmarks used in this area is summarized in Table~\ref{tab:datasets-benchmarks}, offering researchers an organized reference for efficient reasoning evaluation.





\end{document}